\documentclass{article}

\usepackage{amssymb}
\usepackage{mathrsfs}
\usepackage{amsmath}
\usepackage{amsfonts}
\usepackage{xcolor}
\usepackage{graphicx}
\usepackage{bm} 
\usepackage{footnote}
\usepackage{comment}
\usepackage{numprint}
\usepackage{siunitx}
\usepackage{booktabs} 
\usepackage{subcaption}
\usepackage{makecell}
\usepackage{multirow}
\usepackage{array}
\usepackage{booktabs}

\usepackage{PRIMEarxiv}
\usepackage[utf8]{inputenc} 
\usepackage[T1]{fontenc}    
\usepackage{url}            
\usepackage{booktabs}       
\usepackage{amsfonts}       
\usepackage{nicefrac}       
\usepackage{microtype}      
\usepackage{lipsum}
\usepackage{fancyhdr}       
\usepackage{graphicx}       
\usepackage{amssymb}
\usepackage{subcaption}
\usepackage[backend=biber, style=ieee]{biblatex}
\graphicspath{{media/}}     

\title{Heterogeneous Multi-Source Data Fusion Through Input Mapping and Latent Variable Gaussian Process}

\author{
  Yigitcan Comlek $^1$ \\
  Integrated Design Automation Laboratory \\
  Northwestern University \\
  Evanston, IL \\
 \texttt{yigitcancomlek2024@u.northwestern.edu} \\
 \And
  Sandipp Krishnan Ravi $^{1,*}$ \\
  Probabilistic Design and Material Informatics Group \\
  GE Aerospace Research \\
  Niskayuna, NY \\
  \texttt{sandippkrishnan.ravi@ge.com} \\
   \And
  Piyush Pandita \\
  Probabilistic Design and Material Informatics Group \\
  GE Aerospace Research \\
  Niskayuna, NY \\
   \And
  Sayan Ghosh \\
  Probabilistic Design and Material Informatics Group \\
  GE Aerospace Research \\
  Niskayuna, NY \\
   \And
  Liping Wang\\
  Probabilistic Design and Material Informatics Group \\
  GE Aerospace  Research \\
  Niskayuna, NY \\
   \And
    Wei Chen $^*$\\
  Integrated Design Automation Laboratory \\
  Northwestern University \\
  Evanston, IL \\
 \texttt{weichen@northwestern.edu} \\
}

\begin{document}
\maketitle

\def\thefootnote{1}\footnotetext{Joint First Authors}\def\thefootnote{\arabic{footnote}}
\def\thefootnote{*}\footnotetext{Corresponding Author}\def\thefootnote{\arabic{footnote}}

\begin{abstract}
Artificial intelligence and machine learning frameworks have served as computationally efficient mapping between inputs and outputs for engineering problems. These mappings have enabled optimization and analysis routines that have warranted superior designs, ingenious material systems and optimized manufacturing processes. A common occurrence in such modeling endeavors is the existence of multiple source of data, each differentiated by fidelity, operating conditions, experimental conditions, and more. Data fusion frameworks have opened the possibility of combining such differentiated sources into single unified models, enabling improved accuracy and knowledge transfer. However, these frameworks encounter limitations when the different sources are heterogeneous in nature, i.e., not sharing the same input parameter space. These heterogeneous input scenarios can occur when the domains differentiated by complexity, scale, and fidelity require different parametrizations. Towards addressing this void, a heterogeneous multi-source data fusion framework is proposed based on input mapping calibration (IMC) and latent variable Gaussian process (LVGP). In the first stage, the IMC algorithm is utilized to transform the heterogeneous input parameter spaces into a unified reference parameter space. In the second stage, a multi-source data fusion model enabled by LVGP is leveraged to build a single source-aware surrogate model on the transformed reference space. The proposed framework is demonstrated and analyzed on three engineering case studies (design of cantilever beam, design of ellipsoidal void and modeling properties of Ti6Al4V alloy). The results indicate that the proposed framework provides improved predictive accuracy over a single source model and transformed but source unaware model.

\end{abstract}


\section{Introduction}
Artificial intelligence (AI) and machine learning (ML) methods have seen extensive proliferation in the engineering domain. They have enabled the development of surrogates that provide key insights into the system or process under study. They have also served as computationally cheap surrogates that could be further leveraged in optimization tasks or explainable AI frameworks. The incorporation of AI\textbackslash ML for design of engineering systems and processes have pushed the performance envelopes to a greater extent. Since a major class of these methods solely depends on data to identify an accurate map between the input and the output space, their performance fall short for the cases where the cost of data (may it be economical, time or effort) is extremely high. Such a conundrum is more prolific in the engineering community, where the cost of data prevents the development of accurate models which further hinders the optimization and design of the engineering systems. Several branches of AI\textbackslash ML have taken their pass to address this problem. These efforts have varied from probabilistic models which provide uncertainty estimations \cite{ravi2023uncertainty}, white box models \cite{luan2024physics, luan2023physics}, physics informed ML \cite{karniadakis2021physics, cuomo2022scientific}, and transfer learning \cite{krishnan2022transfer,ravi2023probabilistic}. More recently, the domains of multi-source modeling and data fusion have created an impact in driving solution towards efficient  metamodeling in limited data conditions \cite{ravi2024interpretable, yousefpour2024gp+, eweis2022data, foumani2023multi, zanjani2024safeguarding}. They identify and fuse data from different sources (e.g., fidelity, complexity) to provide an accurate model on the source of interest. 

A significant challenge in multi-source modeling endeavors is the variation in the input parameter space of the sources. Various factors may necessitate different parameterizations in the model inputs, leading to heterogeneous data sources. Examples include the increasing complexity of design details, different classes of designs, enhanced fidelities of simulations \cite{menon2024multi}, and varying scales of study, among others. To further illustrate the necessity and impact of this work, consider the design of an aerodynamic blade. The initial iterations and analysis are primarily driven by the efficiency it offers. However, as the blade undergoes design updates, additional details and features are incorporated to address other nuanced functionalities, such as fixture features and locations, manufacturing constraints, etc. In such scenarios, the input parameter space of subsequent design iterations evolves in representation and complexity. Furthermore, in the same blade design endeavor, as the fidelity of the structural simulation increases, the parameterization of the material behavior (e.g., elastic, plastic, composite) and the type of simulation (e.g., linear, non-linear, static, dynamic) can lead to heterogeneity across sources. Such scenarios can also be observed in other domains of engineering, necessitating a robust heterogeneous multi-source data fusion framework.


Numerous approaches have been proposed in the domains of heterogeneous transfer learning and data fusion, as highlighted in references \cite{day2017survey, bao2023survey, shi2010transfer, duan2012learning, wang2011heterogeneous, liu2023learning, gorodetsky2020mfnets, jin2021combining}. Xiaoxiao et al. \cite{shi2010transfer} focused on resolving feature heterogeneity by investigating a common feature space for domains from two sources. Lixin et al. \cite{duan2012learning} introduced a heterogeneous feature augmentation framework, where two different projection matrices are constructed to address heterogeneous features with varying dimensions. This approach transforms the input parameter space from the two domains into a common subspace to measure domain similarity. Additionally, two new feature mapping functions are proposed to augment the transformed input parameter space with the original features. Wang and Mahadevan \cite{wang2011heterogeneous} proposed an approach designed to learn mapping functions that project the domains of two sources into a new latent space. One of the most recent contributions is the development of a heterogeneous multi-task Gaussian Process (GP) by Liu et al. \cite{liu2023learning}, formulating a GP framework with multi-task capabilities for heterogeneous inputs. More relevant to the framework proposed in this research work, prior studies by Tao et al. and Hebbal et al. \cite{tao2019input, hebbal2019multi, hebbal2021multi} have targeted and demonstrated methodologies applicable to engineering design problems. These methods are built upon the concept of space mapping in the domain of electromagnetic system design \cite{bandler1994space, koziel2018implicit, jiang2018space}. Tao et al. \cite{tao2019input} proposed identifying a linear mapping between two input domains by minimizing the error between the projected source and the true source. A more integrated approach was developed by Hebbal et al. \cite{hebbal2019multi, hebbal2021multi}, aimed at finding the mapping between different input parameter spaces and simultaneously mapping the inputs and outputs through deep GP models. The initial layers of the deep GP model are designed to capture the unknown mapping between the two input parameter spaces, while the subsequent layer facilitates the mapping between the common transformed input parameters and the outputs. This setup enables simultaneous training of the mapping between the two input domains and conventional mapping between inputs and outputs. Majority of the prior work in this domain have been focused and demonstrated on two sources. The engineering application of such methods have not been extended to more than two sources differentiated by design complexity and\textbackslash or fidelity.


In this work, a source-aware multi-source data fusion framework capable of accounting for heterogeneous inputs from more than two sources is developed and studied. The developed framework consists of two major stages: (1) Heterogeneous Mapping and (2) Multi-Source Data Fusion. Heterogeneous mapping enables the transformation of different (both overlapping and non-overlapping) input parameter spaces into a common parameter space via Input Mapping Calibration (IMC) \cite{tao2019input}, a linear mapping scheme. Once all mappings are identified with respect to a common reference source, a unified multi-source data fusion model, enabled by the Latent Variable Gaussian Process (LVGP) \cite{ravi2024interpretable}, is trained on these transformed data. The trained LVGP model can make predictions across all sources and offers interpretability through the latent variables and a dissimilarity metric. The latent variables and dissimilarity metric can be leveraged to understand the differences and similarities across all sources of information. The proposed method is demonstrated through three engineering case studies: the design of a cantilever beam across different designs of cross-sections, the design of an ellipsoidal void across varying complexities and fidelities, and the modeling of the performance of Ti6Al4V alloy across different manufacturing processes. Across all case studies, it was observed that the proposed framework performed better compared to conventional methods.

The remainder of the paper is organized as follows. Section 2 articulates the heterogeneous multi-source data fusion model that is based on IMC and LVGP. Section 3 presents the three case studies and in-depth analysis into the input mapping, multi-source data fusion modeling, and comparison with other existing state-of-the-art models. Finally, the paper is concluded through Section 4 with the takeaways and discussions from this research work.

\section{Heterogeneous Multi-Source Data Fusion}

\begin{figure}[h]
      \centering
		\includegraphics[width=1\textwidth] {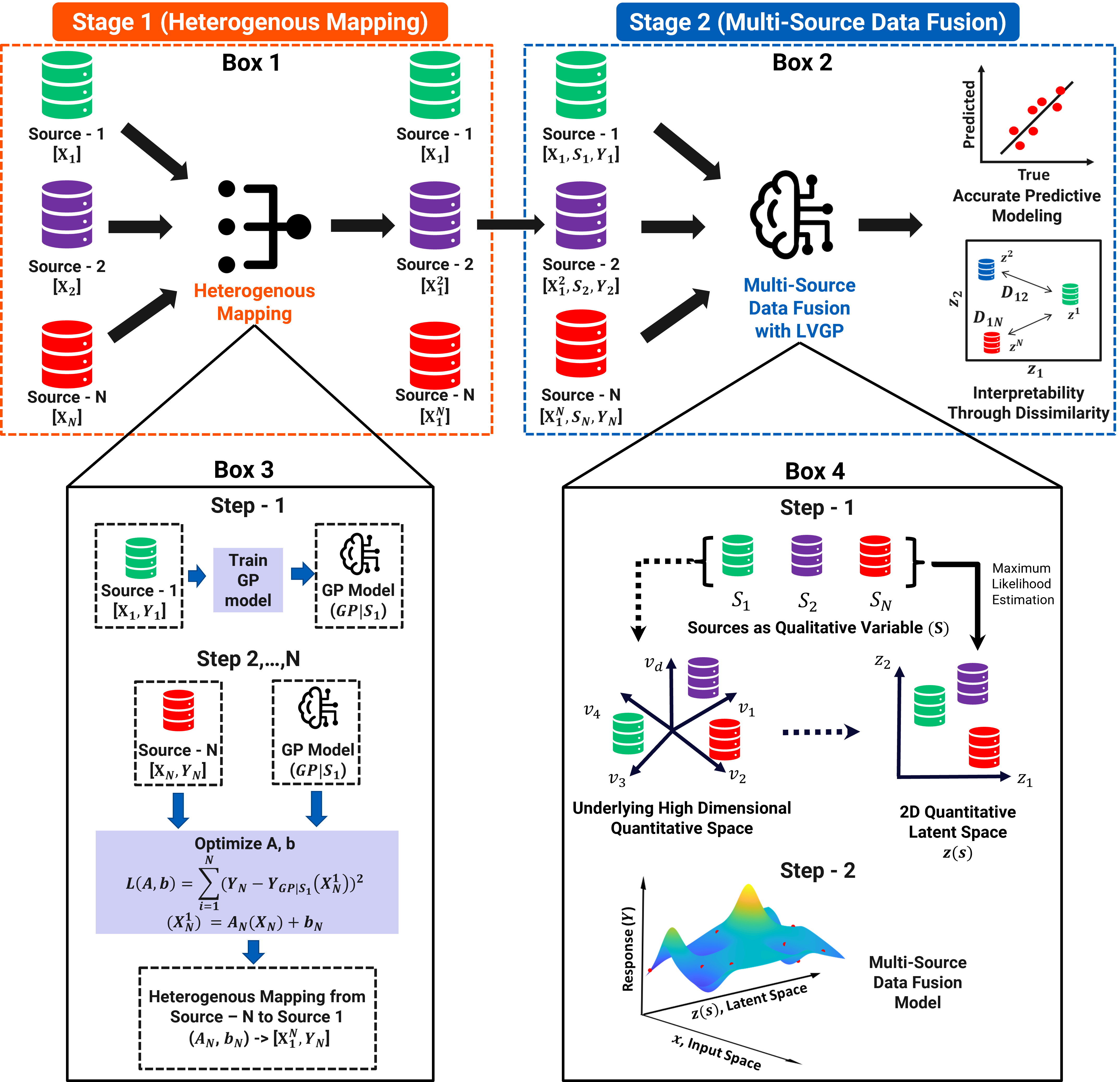}
		\caption{The heterogeneous multi-source data fusion framework}
		\label{fig:Framework}
\end{figure}

In the analysis of engineering systems, valuable insights can be gleaned from various sources of information. However, the inherent diversity in data collection methodologies across these sources often results in discrepancies and underlying differences that are not incorporated into the modeling of engineering systems. To address this challenge, previously a novel approach leveraging a multi-source data fusion technique with LVGP \cite{ravi2024interpretable} was proposed. This technique amalgamates information from disparate sources into a unified, source-aware model, accounting for not only inter-source relationships but also known and unknown underlying physical parameters associated with each source of information. On the other hand, this approach is limited to cases where the sources of information share the same input domains. As previously stated, many engineering applications share sources of information with varying (heterogeneous) input domains. To overcome this challenge, a heterogeneous multi-source data fusion method is proposed. The framework consists of two stages, Heterogeneous Mapping (Stage 1) through IMC, and Multi-Source Data Fusion (Stage 2) through LVGP, as shown in Figure \ref{fig:Framework}. The details of each stage along with brief reviews of the methods (IMC and LVGP) are provided below. 

\subsection{Stage 1: Heterogeneous Mapping}

In the first stage, the data from available sources is collected and IMC is implemented to map the input space of the sources into a reference source domain (Box 1 in Figure \ref{fig:Framework}). A review of the IMC method and it's implementation on the proposed method is provided below.

\subsubsection{Review of Input Mapping Calibration} 
Initially developed for accounting different high-fidelity and low-fidelity model inputs for multi-fidelity applications, IMC is a calibration method that provides a single input domain that is mapped from different input domains \cite{taoimc2019}. For multi-source data fusion, IMC can be used for modeling different sources of information that share the same output but contain different input domains. 

For a given set of two sources of information, Source 1 and Source 2, with their respective different input ($x_1,x_2$) with dimensions ($d_1, d_2$), and output ($y_1(x_1), y_2(x_2)$), the mapping is achieved by transforming one source's inputs into the other source's (reference source) domain. Usually, the reference source is selected as the source with the highest amount of data samples to achieve the highest accuracy during the mapping. For illustration purposes, Source 2 is mapped onto Source 1. The mapping is done through a parameterized calibration function, denoted as $g(x,\beta)$ where $x$ is the input domain and $\beta$ is the parameters to be estimated. The parameters are estimated by minimizing the difference between the output domains of the sources. By setting $x_1 = g(x_2,\beta)$, the outputs can be matched through 

\begin{equation}\label{eq:imc1}
      y_2(x_2)=y_1(g(x_2;\beta)) + \epsilon
\end{equation}

where $\epsilon$ is assumed to be an independently and identically distributed error. For the cases where no prior information exists, the form of the calibration function $g(x,\beta)$ can be formulated as a full-linear transformation

\begin{equation}\label{eq:imc2}
      g(x; A, b) = Ax + b
\end{equation}
 
where $A$ is $d_1 \times d_2 $ matrix and $b$ is a $d_1 \times 1$ vector under the parameter vector $\beta$.  The parameters are estimated by minimizing the mean squared loss ($L$) between the outputs of the sources as shown in Equation \ref{eq:imc3}. A standard normalization is performed on both the input spaces prior to optimization for ease of convergence. Throughout the paper, this transformation is indicated in the equation through the $Norm$ subscript.

\begin{equation}\label{eq:imc3}
      min(L) = (y_2(x_2)-y_1(g(x_2;\beta)))^2
\end{equation}

It is important to note that the transformation between the two input spaces does not necessarily have to be linear. For the current work, a linear transformation is chosen to balance the robustness and versatility of the proposed method with accuracy and performance. The transformation matrix between the input parameter spaces have to be solved element-wise through an optimization algorithm due to the non-linearity introduced from the model in the loss function. The possibility of solving the transformation matrix elegantly through decomposition and rank reduction (as done in transfer component analysis \cite{pan2010domain}) might be limited. To balance the computational cost, highly non-uniqueness of the solutions and limited data, a linear transformation is chosen. Alternatives for the mapping include non-linear or a kernel-based transformations. Non-linear transformation can be attractive alternative if appropriate caution is taken to address some of the issues that arise from solving a large dimensional optimization problem. The prospect of linear transformation might be limiting in terms of transformation achieved and accuracy. However, the authors hypothesize that the multi-source data fusion through LVGP can balance the perceived limitation by learning the relationships between the sources through latent variables and using this knowledge for better predictive modeling. The details of latent variable learning is provided in Section \ref{lvgp}.



\subsubsection{Multi-Source Mapping}
To implement IMC for multi-source data fusion modeling, the reference source data is selected as the source with most amount of available information (data). For demonstration purposes, this source is assigned as Source 1 in Figure \ref{fig:Framework} Box 3. Heterogeneous mapping is achieved in multiple steps, depending on the number of available sources. In the very first step, a GP model is built on the reference source data. This GP model, denoted as $GP|S_1$, is used as a surrogate to estimate the response of the system based on the calibrated input domain for each of the remaining sources. In the second step, the input domain of Source 2 is calibrated in terms of Source 1 domain through Equation \ref{eq:imc1}, while using the calibration function (Equation \ref{eq:imc2}) with parameters to be estimated. The parameters are optimized by matching the outputs of two sources and minimizing the loss function (Equation \ref{eq:imc3}). The loss function is minimized using a Genetic Algorithm approach. This step is repeated for each of the remaining sources until all source input domains are mapped and calibrated to the reference source domain. Finally, the modified input space consisting of the inputs from the reference source and mapped inputs from other sources are passed onto Stage 2 for multi-source data fusion.

\subsection{Stage 2: Multi-Source Data Fusion}

With its demonstrated accurate and interpretable modeling capabilities, LVGP is employed for building a source-aware multi-source data fusion model. Specifically, LVGP model is built over the mapped source domains for the second stage. A review of the LVGP along with the details of multi-source data fusion through LVGP is provided below.

\subsubsection{Review of Latent Variable Gaussian Process}\label{lvgp}


One of the main contribution of this paper lies in the modeling of the multiple sources of information as a representative qualitative (categorical) input in addition to known quantitative (numerical) mapped inputs through the novel LVGP ML model. LVGP was initially developed for mixed-variable modeling \cite{zhang2020latent,wang2020featureless,iyer2020data,wang2021data,prabhune2023design,comlek2023rapid} and has been recently extended to integrate information from different sources \cite{ravi2024interpretable} . 

Consider an input space, $\mathbf{b} = [\mathbf{x^T},\mathbf{t^T}]$ where both quantitative, $\mathbf{x}=[x_1,x_2,..x_m] \epsilon R^m$, and qualitative $\mathbf{t}=[t_1,t_2,...,t_q] \epsilon R^q$ input variables are included in this new domain. Each qualitative variable can have $j$ number of options (levels), i.e $t_i = [l_1(t_i),l_2(t_i),...,l_j(t_i)]$ where $i = 1,2,...,q$. For a given response of interest, $y(b)$, the influence of each qualitative variable can be represented by an underlying quantitative (numerical) space, $v_{t_i}(l_j) = [v_1(l_1),v_2(l_1),...,v_k(l_j)] \epsilon R^k$. This underlying quantitative space can possibly be undiscovered, unknown, complex, or high-dimensional. As a result, the main idea of LVGP lies in learning a low-dimensional  quantitative latent space to approximate the original underlying space through statistical inference using the available input-output data. According to the original LVGP paper, a two-dimensional latent variable vector is sufficient to express the influence of the qualitative variables on the response of interest  \cite{zhang2020latent}. Therefore, qualitative variable, $t_i$ can be represented as latent variables $\mathbf{z}(t_i) = [z_1(l_1),z_2(l_1),z_1(l_2),z_2(l_2),...,z_1(l_j),z_2(l_j)]$, where each level $l_j(t_i)$ is represented with a 2D latent vector $[z_1(l_j),z_2(l_j)]$ . Thus, the previously defined mixed-variable input space, $\mathbf{b} = [\mathbf{x}^{T},\mathbf{t}^{T}]$, then becomes $\mathbf{w} = [\mathbf{x}^{T},\mathbf{z(t)}{^T}] \epsilon R^{m + 2 \times q}$. With the newly defined input space, the  GP correlation function is denoted as,

\begin{equation}\label{eq:corr2}
        c(w,w') = exp(-\sum_{i =1}^{m} (\phi_{i}(x_{i}-x_{i}^{'})^2) - \sum_{j =1}^{q} ||z_{1,j} - z_{1,j}^{'}||^{2}_2 + ||z_{2,j} - z_{2,j}^{'}||^{2}_2)
\end{equation}

Consequently, the LVGP model parameters $\mu$, $\sigma^2$, and $\phi$, along with the latent variables, $\mathbf{z}$, are estimated through maximum likelihood estimation (MLE) of the log-likelihood function

\begin{equation}\label{eq:likelihood}
        l(\mu,\sigma,\phi,\mathbf{z}) = -\frac{n}{2} ln(2\pi\sigma^2) - \frac{1}{2}ln|C(\mathbf{\phi},\mathbf{z})| - \frac{1}{2\sigma^2}(\mathbf{y}-\mu\mathbf{1})^{T}C(\mathbf{\phi},\mathbf{z})^{-1}(\mathbf{y}-\mu\mathbf{1})
\end{equation}

where $n$ is the number of available training samples, $C$ is a $n \times n$ correlation matrix with $C_{ij} = c(w_i,w_j)$ for $i,j = 1,2,...,n$, $\mathbf{y} = [y_1,y_2,...,y_n]$ is the observed response vector, $\mathbf{1}$ is a vector of ones with a size of $n \times 1$. Once all the model parameters are estimated, the LVGP model can make predictions, on any given input with the quantified prediction uncertainty. 

By incorporating information sources as qualitative variables and mapping them to latent variables, this approach provides physics-based dimension reduction that could potentially explain the "cause-effect" relationship between the inputs (information sources) and outputs (response). Analyzing the spatial locations of the latent variables (sources) in the latent space cam provide physical insights into the information sources. For instance, Variables that are closer together in the latent space can have a similar impact on the response. Moreover, in scenarios where no prior knowledge exists regarding information sources, the LVGP model can help extract and reveal the relationships between the sources. To understand the underlying similarities and differences between information sources, a dissimilarity metric ($D$) is introduced. This metric quantifies the dissimilarities between the sources through a distance metric, measured by the Euclidean distance of the data source ($z_j$) from the reference data source ($z^*$) in the latent space (Figure \ref{fig:Framework} Box 2). The metric, normalized by the maximum latent distance, is given in Equation \ref{eq:D_1}. In the LVGP model, the latent variables are constrained to be between $z = [-3, 3]$ based on typical practice in previous applications \cite{zhang2020latent,wang2020featureless,iyer2020data,wang2021data,prabhune2023design,comlek2023rapid}. The reference data source will be placed at $z^*= [0, 0]$ in the latent space. Consequently, the most dissimilar information source from the reference source is going to be placed at $[±3, ±3]$, making $z_{max} = [±3, ±3]$. As a result, the maximum distance between the reference and given information source can be $||z_{max} - z^*||= 3\sqrt{2}$. Therefore, the introduction of LVGP enables the development of a a source-aware (categorical tagging) and interpretable (latent variable analysis, dissimilarity metric) data fusion model. Further details of the multi-source data fusion modeling are provided in the subsequent section.


\begin{equation}\label{eq:D_1}
D(z^j) =  \frac{||z^j-z^*||}{||z^{max}-z^*||} \longrightarrow \frac{||z^j-z^*||}{3\sqrt{2}}
\end{equation}


\subsubsection{Multi-Source Data Fusion Through LVGP}

To incorporate sources into the modeling, a new input domain is created $[X,S]$. Here, $X$ represents the common mapped numerical parameters shared across sources, while $S$ denotes a newly introduced categorical variable representing the distinct sources along with their respective outputs $[Y]$ (Figure \ref{fig:Framework}, Box 2). Once the data is collected, the LVGP model is built in two steps as shown in Figure \ref{fig:Framework}, Box 4. In the first step, sources ($[S]$) are treated as a qualitative variable and integrated into the multi-source model using statistical inference techniques similar to those used for qualitative variable $t$ in Section \ref{lvgp}. Each information source is labeled with a unique identifier ($l$ in Section \ref{lvgp}), and the underlying quantitative space of each source is mapped into a 2D quantitative latent space. The latent variables ($z(s)$) are estimated through MLE of the Equation \ref{eq:likelihood} during the LVGP model training. Consequently, the second step provides the trained multi-source data fusion model based on input variables ($X$), the source variable ($z(s)$), and the response ($Y$). Finally, the trained model can make predictions for any given source input that contains it's mapped numerical parameters ($X$) and the assigned source variable ($S$).


\section{Case Studies}

The proposed framework is demonstrated through three case studies based on engineering applications. Each case study aims to showcase the the greater efficacy of the proposed approach for different scenarios. The first case study, Section \ref{beam}, demonstrates a scenario where sources of data differ across designs due to different parametrizations. This study is shown for cantilever beam with different cross-sections. The second case study, Section \ref{void}, shows a scenario where the sources of data differ in design complexity (2D \& 3D), and fidelity (elastic \& plastic). This study is shown for an ellipsoidal void sources with varying dimensionality (complexity) and elastic/plastic structural analysis (fidelity). Finally, the third case study, Section \ref{void}, displays a scenario where sources of data are from different manufacturing modalities of the same material system. Specifically, this study was shown on a titanium alloy generated by different additive manufacturing processes that share no common input parameters.   

Four important contributions of the work is demonstrated through the case studies. First, the LVGP approach is flexible to handle many multi-source scenarios with varying conditions such as complexity, fidelity, and modality. Second, predictive capabilities of multi-source data fusion through IMC and LVGP exceeds the state-of-the-art IMC with GP. Third, in addition to providing better predictions, the LVGP approach also provides interpretability through the learned latent spaces of the sources. Fourth, for the scenarios where a source of interest does not have high amount of available data, the proposed method provides better predictive capabilities compared to a model built based on that specific source. Specifically, it is demonstrated that for the source with low amount of data, the multi-source LVGP with mapped inputs offers more accurate predictions compared to a single-source GP model built on the original input space. For predictive accuracy comparisons between different models, normalized root mean squared error metric, abbreviated as NRMSE, was used. 


\subsection{Case Study 1: Cantilever Beam Design} \label{beam}

\begin{figure}[h!]
      \centering
		\includegraphics[width=0.9\textwidth]{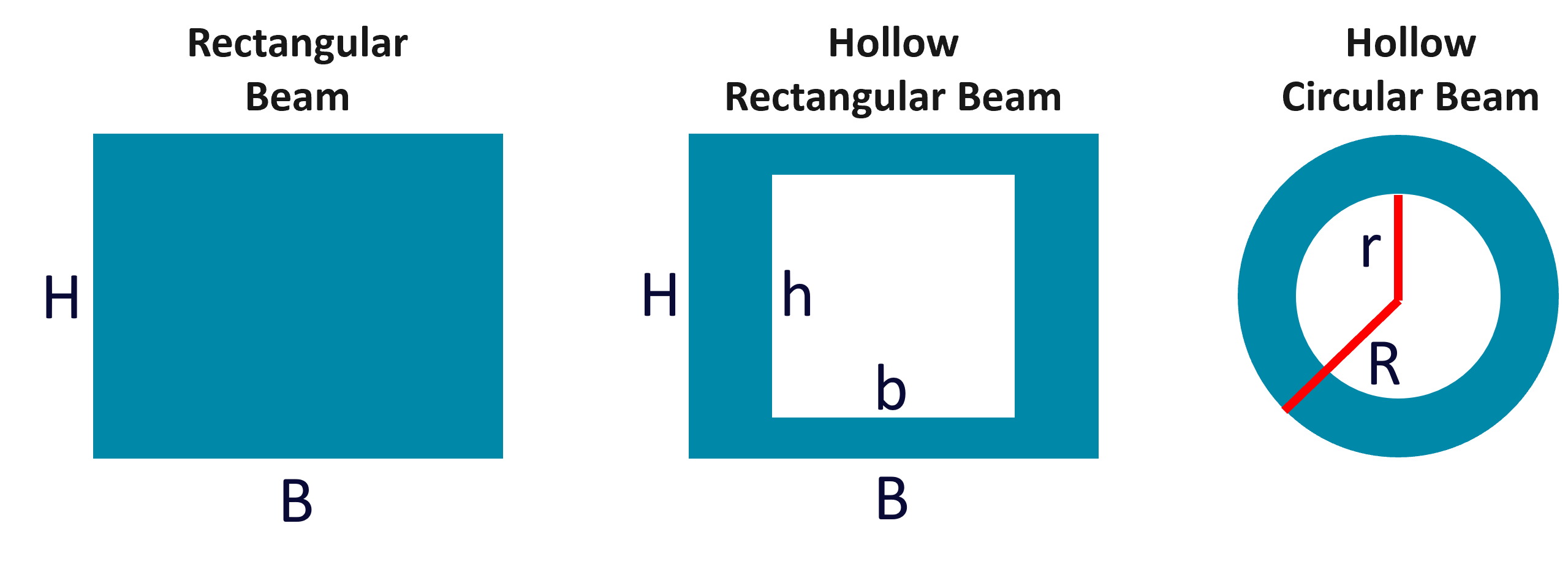}
		\caption{Different cantilever beam designs (sources) with varying parametrization}
		\label{fig:Beam}
\end{figure}

The cantilever beam design is a well-known problem of interest for many engineering applications such as bridges, towers and buildings. A cantilever beam is an horizontally elongated structure that is fixed on one end. Typically, a structural load is applied on the unsupported end of the beam to observe it's deflections and support capabilities. In the context of engineering design, the cantilever beam can take many shapes and forms for optimized performance. Common designs (sources) of the beam include Rectangular Beams, Hollow Rectangular Beams, and Hollow Circular Beams with varying design parameters as depicted in Figure \ref{fig:Beam}. Conventionally, each of the beam design would be considered individually for modeling and optimization purposes as the input domains differ. For instance, the rectangular beam contains height (H) and length (B) variables, whereas the hollow beam is constructed from two radii (R, r) variables. This poses a challenge to incorporate multiple domains of beam into a single model for further design modeling. 

\subsubsection{Data Description}
For each of the sources, data that consists of the respective design variables and the common output, maximum deflection ($d$) of the beam is created. The three sources considered for this study is depicted in Figure \ref{fig:Beam}. For uniform notation, Rectangular Beam, Hollow Rectangular Beam, and Hollow Circular Beam sources are named RB, HRB, and HCB, respectively. To demonstrate the advantages of the proposed method, RB is selected as the source with the most data available, making it the reference source for mapping. Similarly, it is assumed that the HCB as the source with least amount of data. Hypothetically, this source can be considered as an expensive to obtain data from or simply a source where amount of information available is limited. As a result, the input domains of HCB and HRB are mapped to the input domain of RB. The details of the dataset used for this study are given in Table \ref{tab:beam_data}.

\begin{table*}[ht!]
\small
\centering
\caption{Cantilever beam data description\label{tab:beam_data}}
\renewcommand{\arraystretch}{1.3}
\begin{tabular}{>{\centering\arraybackslash}p{3cm} >{\centering\arraybackslash}p{3cm} >{\centering\arraybackslash}p{2cm} >{\centering\arraybackslash}p{3cm} >{\centering\arraybackslash}p{3cm}}
\toprule
Source Type & Input Variables &Output Variable& Number of Training Samples  & Number of Testing Samples \\
\toprule
Rectangular Beam (RB) & $B (m),H (m)$ & & $30$  &  $1000$\\
\\
Hollow Rectangular Beam (HRB) & $B(m),H (m),$ $b(m),h(m)$ & Maximum Deflection ($m$)& $25$ & $1000$\\
\\
Hollow Circular Beam (HCB) & $R (m),r (m)$ & &$8$& $1000$\\
\toprule
\end{tabular}
\end{table*}


\subsubsection{Results}
\paragraph{Heterogeneous Mapping} 
The first step of the proposed method is to map different domains of sources into a single domain. Supported with the most amount of data, a GP model on the RB source is built. Based on this GP model, IMC method was implemented to transform the domains of two other source into the domain of RB. The results of the linear transformation are displayed in Equations (\ref{eqn:beam_t1}) and (\ref{eqn:beam_t2}) for HRB and HCB, respectively. For the transformation of HRB to RB, it can be observed that majority of the contribution to $B$ in the transformed is coming from $H$ (with a factor of $1.51$), $B$ (with a factor of $0.26$) and shift of 1.08 (from $b$). The trend remains same for transformed $H$. The majority of the contribution are derived from $B$ (with a factor of $0.10$) and $H$ (with a factor of $0.30$). The shift from $b$ is observed to be minimal. The equivalent $B$ and $H$ for CHB is primarily driven by $R$ and a shift parameter from $b$ as shown through Equation (\ref{eqn:beam_t2}).

\begin{equation}\label{eqn:beam_t1}
\begin{bmatrix} B \\ H \end{bmatrix}_{Norm} = \begin{bmatrix} 0.26 & 1.51 & 0.07 & 0.01 \\ 0.10 & 0.30 & 0.01 & 0.01 \end{bmatrix}  \begin{bmatrix} B \\ H \\ b \\ h \end{bmatrix}_{Norm} + \begin{bmatrix} 1.08 \\ 0.07 \end{bmatrix} \\
\end{equation}

\begin{equation}\label{eqn:beam_t2}
\begin{bmatrix} B \\ H \end{bmatrix}_{Norm} = \begin{bmatrix} 2.00 & 0.00 \\ 0.89 & 0.00  \end{bmatrix}  \begin{bmatrix} R \\ r  \end{bmatrix}_{Norm} + \begin{bmatrix} 0.55 \\ 0.67 \end{bmatrix} \\
\end{equation}

\begin{figure}[h]
      \centering
	   \begin{subfigure}{0.45\linewidth}
		\includegraphics[width=\linewidth]{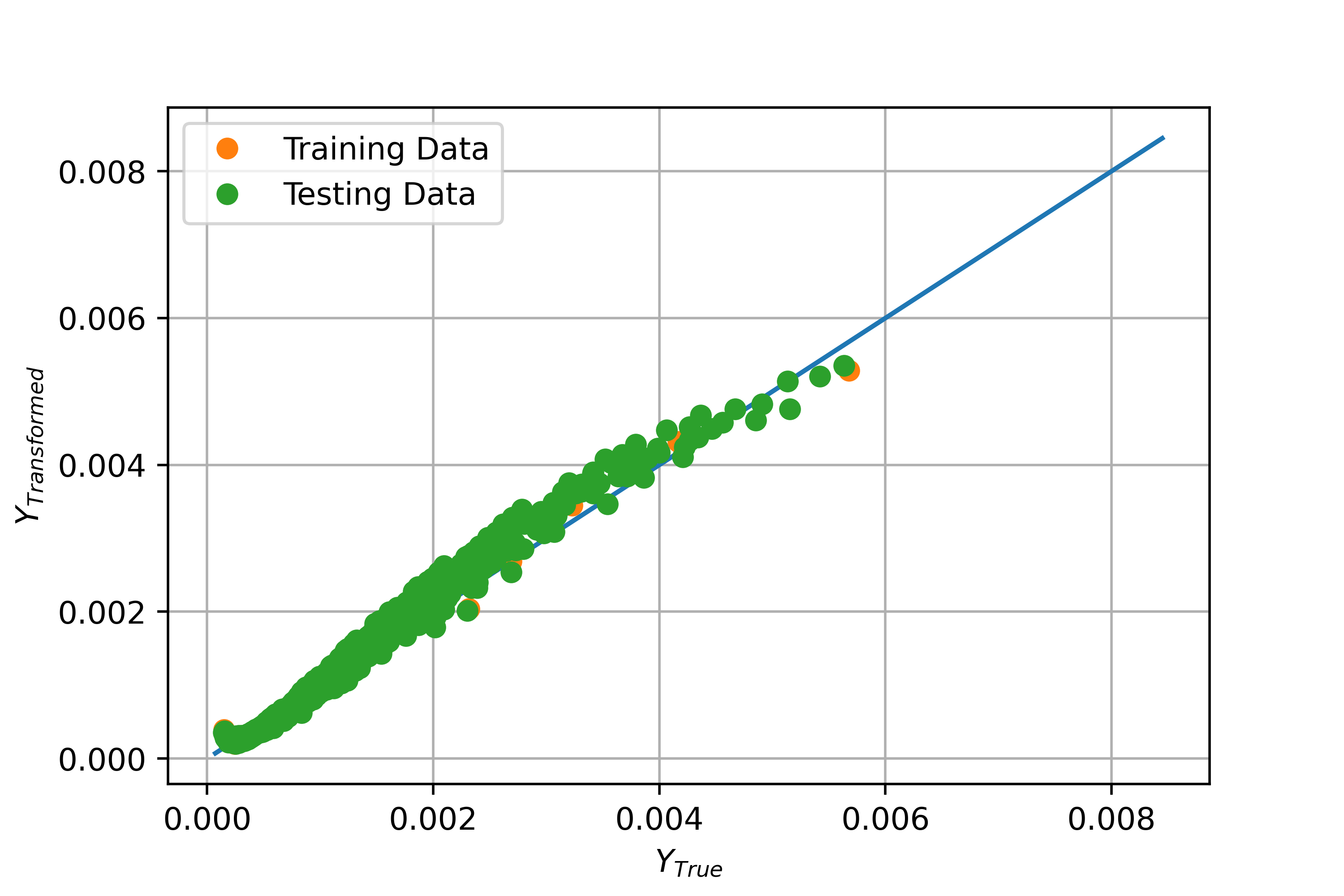}
		\caption{}
		\label{fig:subfig1}
	   \end{subfigure}
	   \begin{subfigure}{0.45\linewidth}
		\includegraphics[width=\linewidth]{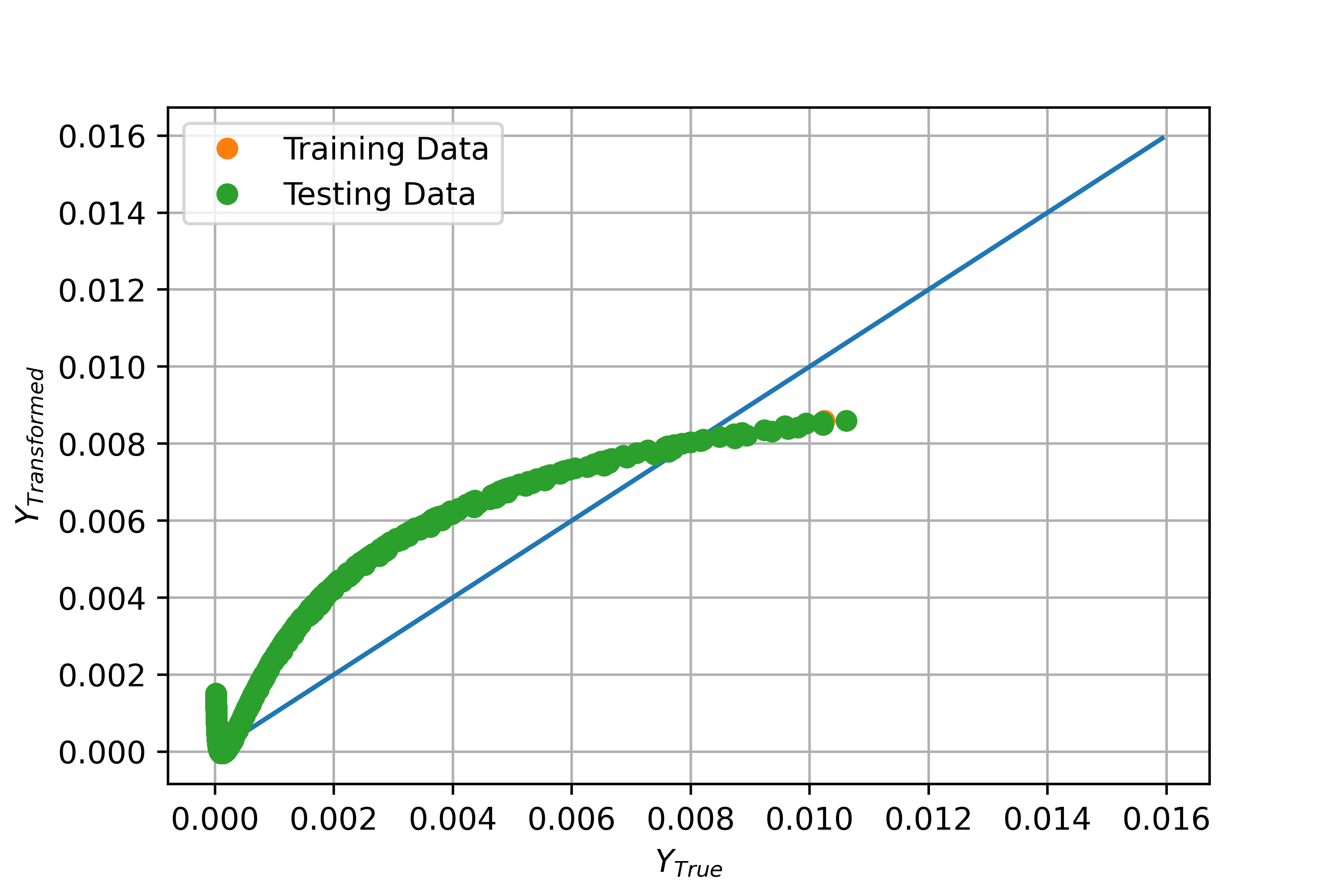}
		\caption{}
		\label{fig:subfig2}
	    \end{subfigure}
	\caption{Predictions based on mapped (a) Hollow Rectangular and (b) Hollow Circular beams using the reference GP model built on Rectangular Beam source}
	\label{fig:beam_transform}
\end{figure}

Once the input domains are mapped, the reference GP model on RB is used to predict the responses of two sources with mapped inputs as shown in Figure \ref{fig:beam_transform}. It is observed that the mapped domain can capture HRB outputs well (Figure \ref{fig:beam_transform}a). On the other hand, acceptable accuracy was observed for the mapped CHB domain (Figure \ref{fig:beam_transform}b). This result may be attributed to the limited amount of available training data or the linear transformation applied. Next, with the newly generated mapped space, the multi-source data fusion models are built.


\begin{figure}[h]
      \centering
	   \begin{subfigure}{0.45\linewidth}
		\includegraphics[width=\linewidth]{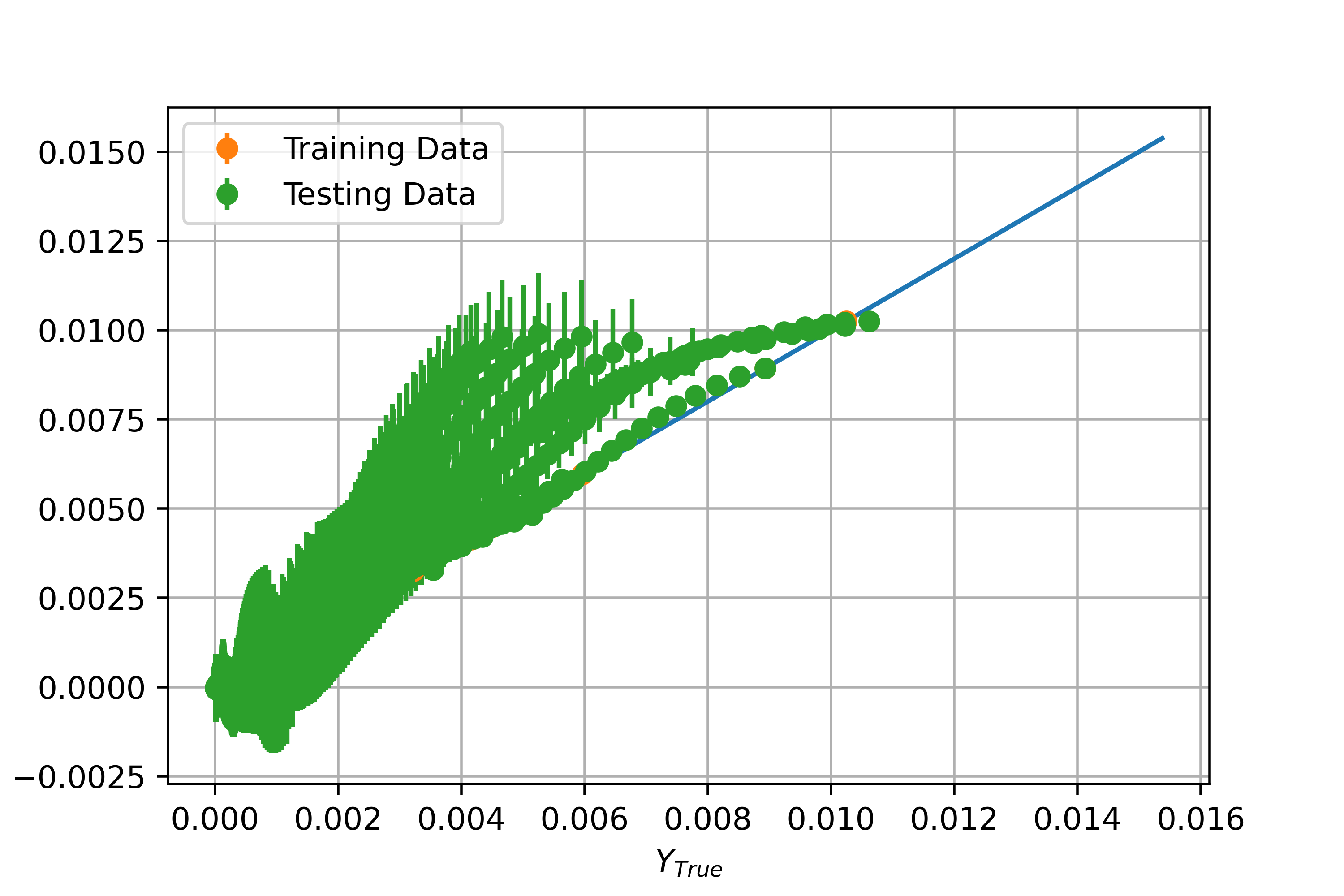}
		\caption{}
		\label{fig:subfig2}
	    \end{subfigure}
	     \begin{subfigure}{0.45\linewidth}
		 \includegraphics[width=\linewidth]{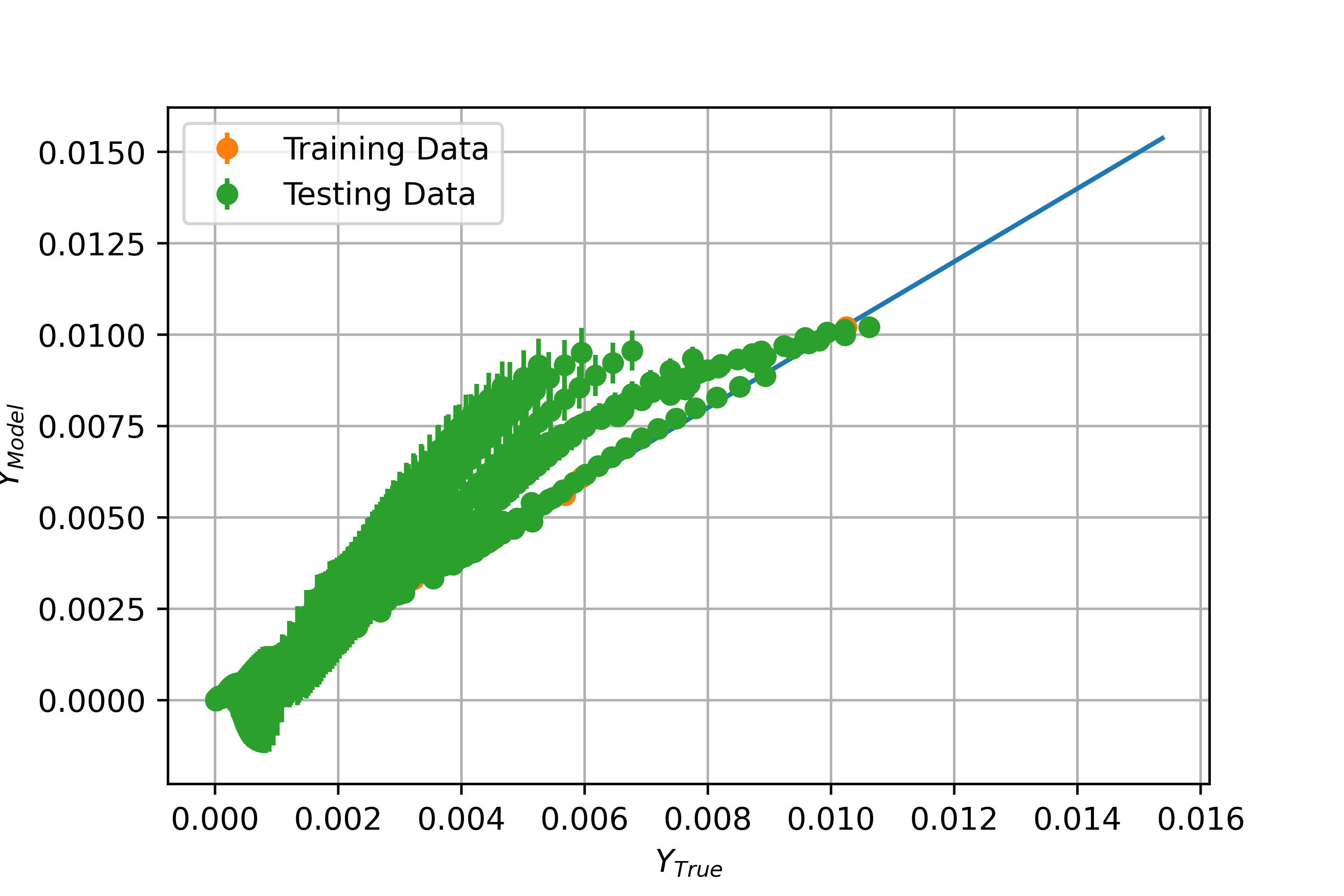}
		 \caption{}
		 \label{fig:subfig3}
	      \end{subfigure}
	      \vfill
	     \begin{subfigure}{0.33\linewidth}
		\includegraphics[width=\linewidth]{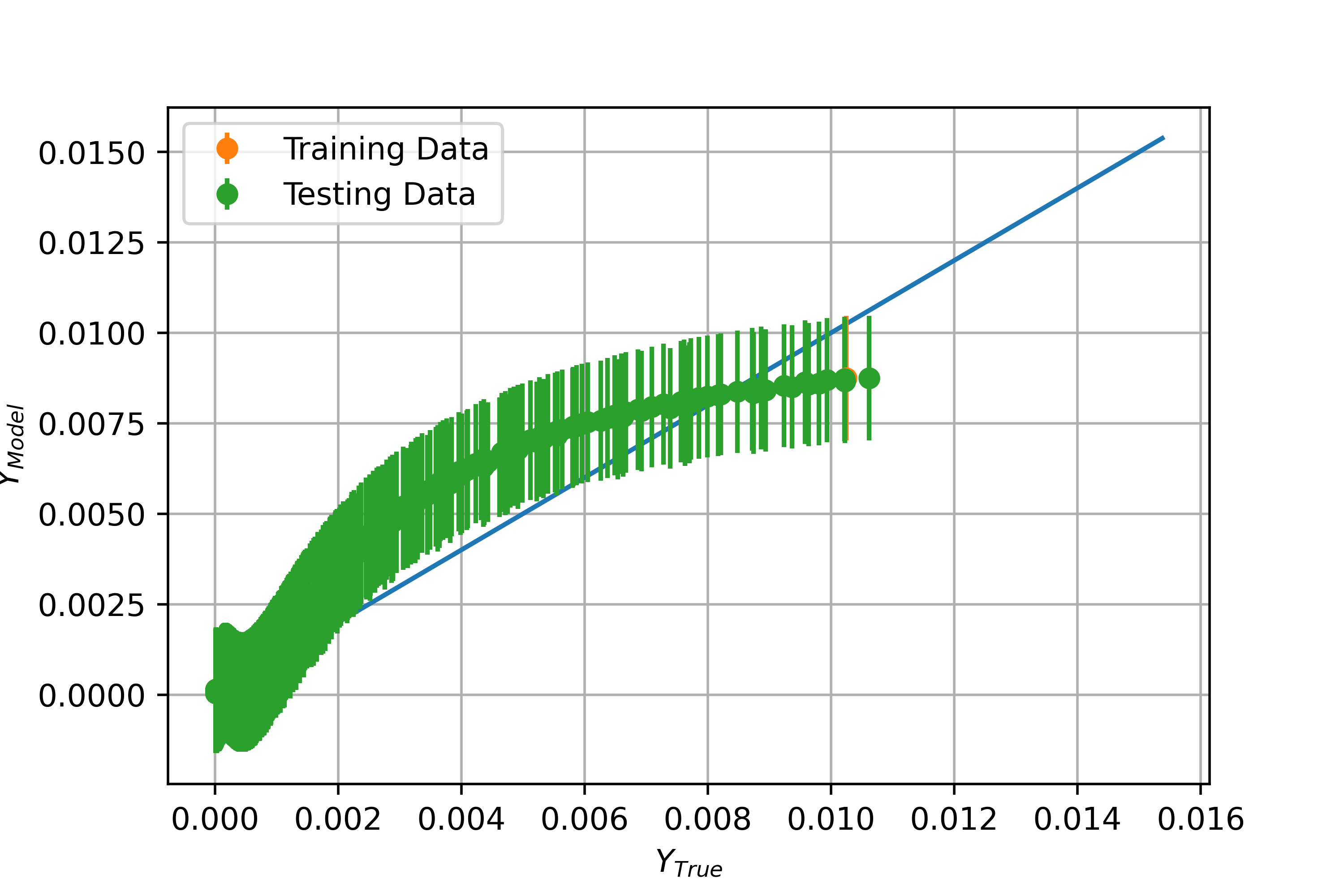}
		\caption{}
		\label{fig:subfig1}
	   \end{subfigure}
	   	     \begin{subfigure}{0.33\linewidth}
		\includegraphics[width=\linewidth]{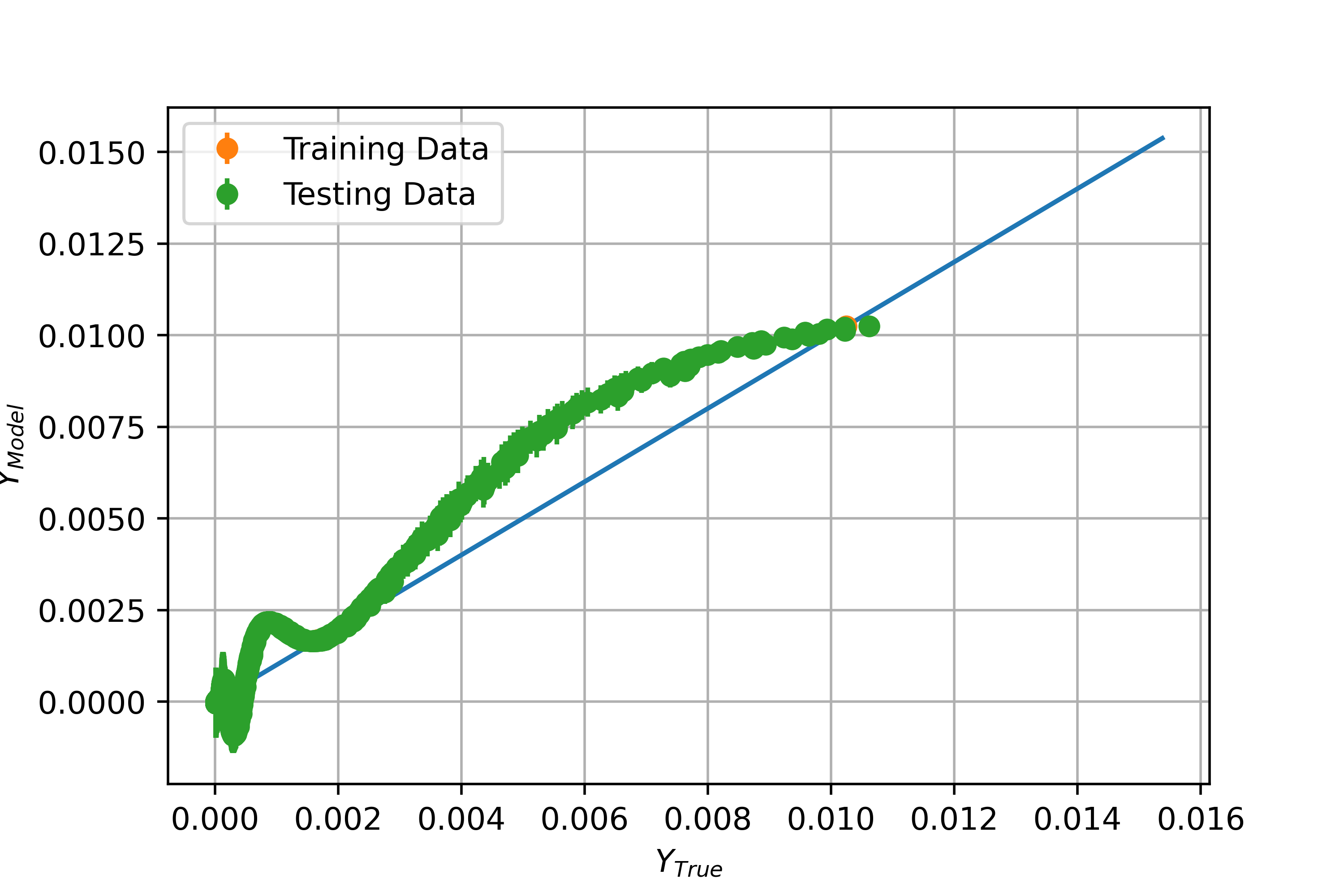}
		\caption{}
		\label{fig:subfig1}
	   \end{subfigure}
	   	     \begin{subfigure}{0.33\linewidth}
		\includegraphics[width=\linewidth]{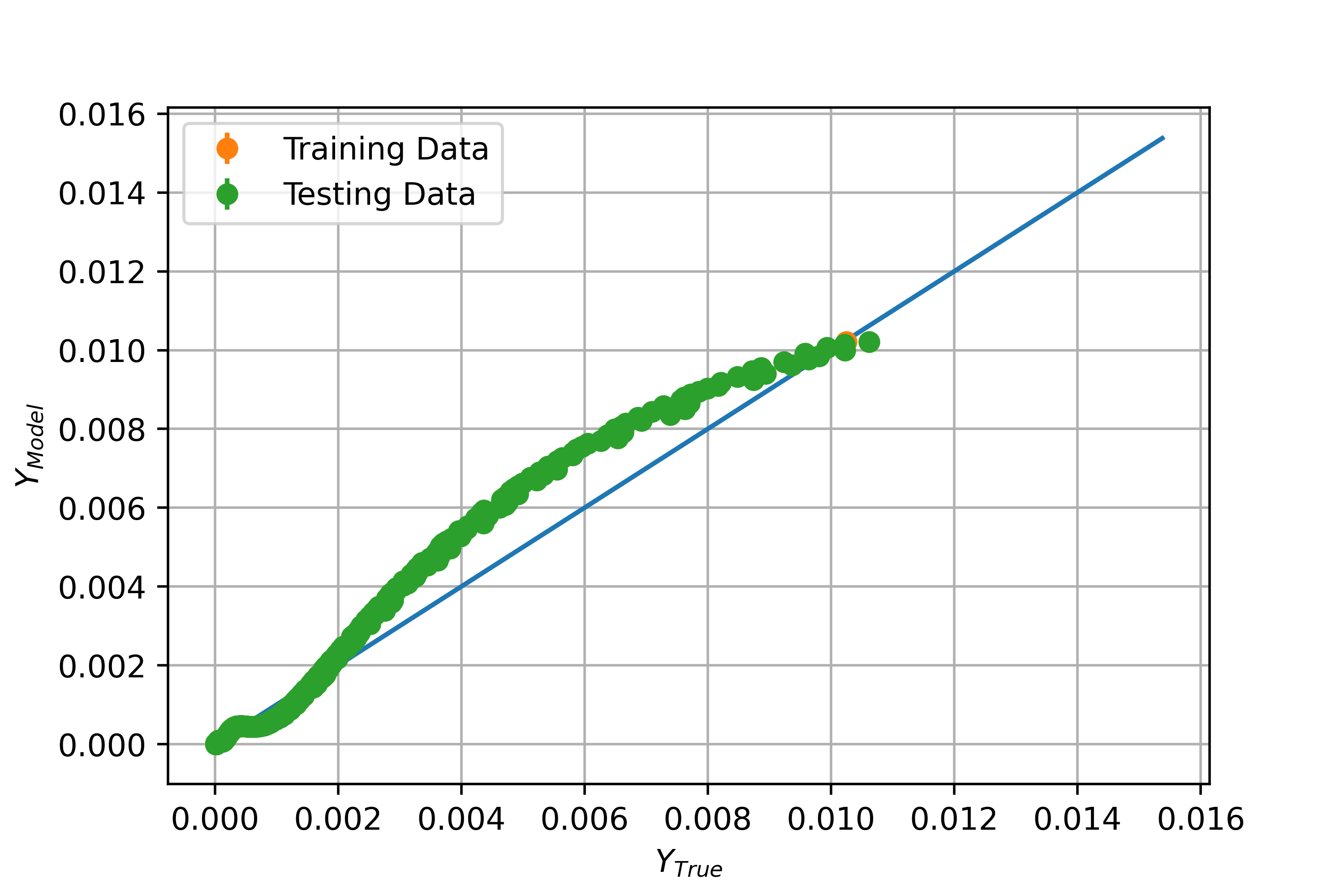}
		\caption{}
		\label{fig:subfig1}
	   \end{subfigure}
	   \vfill
	       \begin{subfigure}{0.45\linewidth}
		  \includegraphics[width=\linewidth]{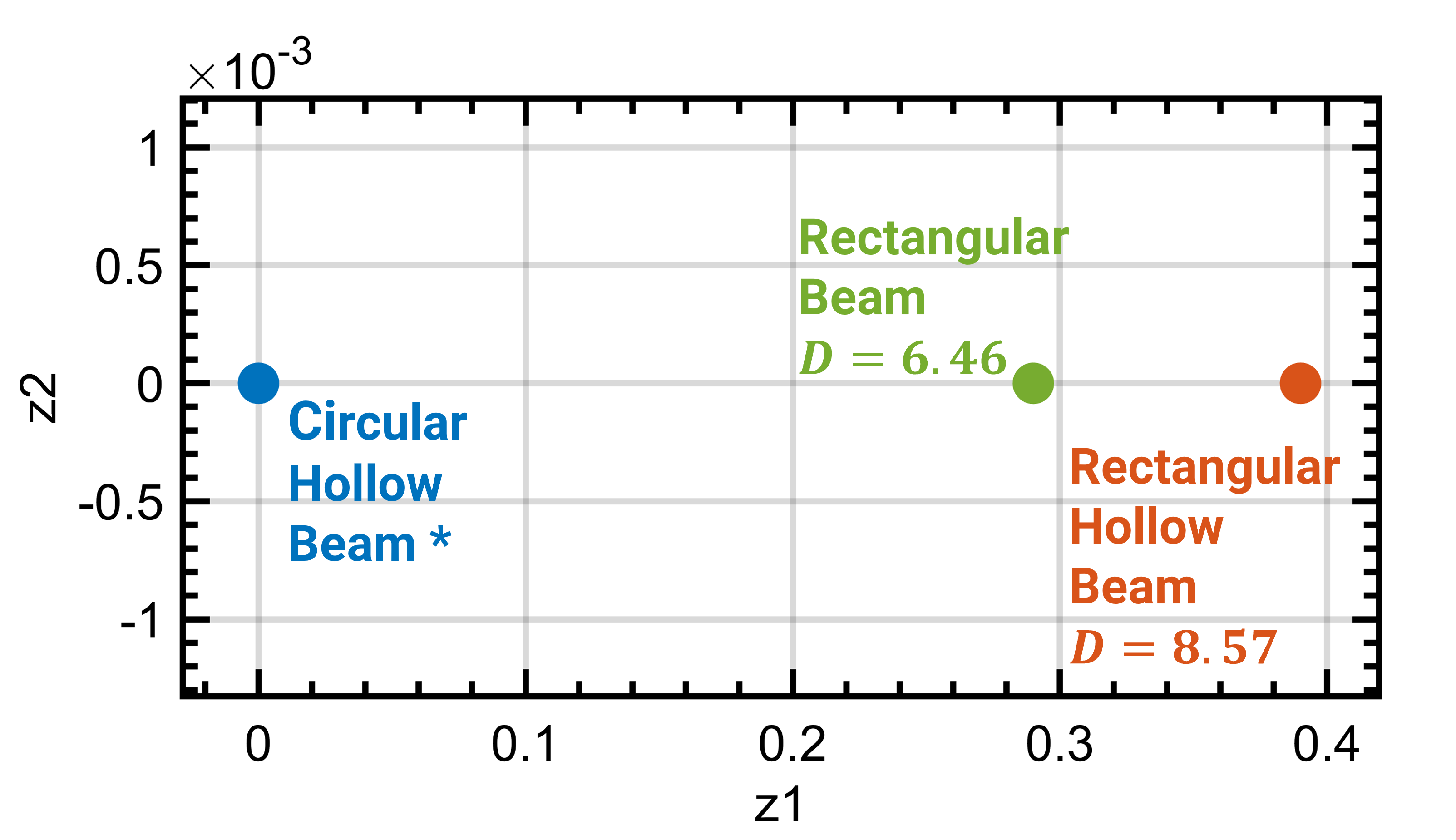}
		  \caption{}
		  \label{fig:subfig4}
	       \end{subfigure}
	\caption{The results of the cantilever beam design study. Predictions on (a) all sources using heterogeneous multi-source GP, (b)  sources using heterogeneous multi-source LVGP, (c) hollow circular beam using single-source GP built on original input space, (d) hollow circular beam using heterogeneous multi-source GP, (e) hollow circular beam using heterogeneous multi-source LVGP. (f) The latent space obtained by the LVGP model}
	\label{fig:beam_summary}
\end{figure}

\paragraph{Multi-Source Modeling Comparison}
The multi-source source-aware LVGP is first compared with multi-source source unaware GP, the current practice with IMC. Both models are trained on inputs from original domain of RB and the mapped domains of HRB and HCB. Specifically, two scenarios are investigated. First, predictions on the training and testing sets on all three sources are compared to evaluate which modeling approach leads to a better and generic model for all sources. The results are provided through Table \ref{tab:beam_pred}, where it is observed that LVGP provides lower NRMSE values for both training (reduction of $87.5\%$) and testing data (reduction of $42.4\%$), indicating that it is a more generalizable predictive model for all the sources. Figure \ref{fig:beam_summary}a\&b shows the parity plots of multi-source GP and LVGP models. It is observed that in addition to providing better predictions, LVGP provides predictions with lower uncertainty, which can be inferred from the error bars. Second, predictions results only on the HCB source are compared as this is the source with the lowest amount of data. Here, a significant improvement in the NRMSE (of $44.6\%$) is observed through LVGP because the source-aware LVGP can make source-specific predictions whereas a regular GP model is unable differentiate between sources and provide a general prediction for the data it's given. The authors hypothesize that one of the contributing factors for such high errors in the source unaware GP is the transformation fit of CHB. The deviations introduced from the transformation (as shown in Figure \ref{fig:beam_transform}b) have no mechanism to be corrected for source unaware GP model. However, multi-source modeling through LVGP differentiates itself with it's ability to account for such deviations evidenced further by the reduced errors.

A natural question to ask is whether it will be more suitable to build a single GP model solely based on CHB source with original input domain. To further demonstrate additional benefits of the proposed method, a single source GP was built on the original HCB data is compared with the multi-source LVGP. The purpose of this comparison is to observe whether the transferred and incorporated knowledge from other sources provide better predictive capabilities compared to a GP model built on the original (non-mapped) input space on that source only. For ease of notation, the GP model is denoted as GP-HCB. NRMSE metrics in Table \ref{tab:beam_pred} demonstrate that LVGP is able to provide more accurate predictions compared to the GP-HCB model (by $52.3\%$). Although the input space of HCB source is mapped to a different domain in the LVGP, incorporating and transferring knowledge from other sources for predictions helps the model make better prediction. This result can be highly beneficial for scenarios where a source of interest has limited amount of information available. For the current case, only 8 training is only made available in the source of interest. Compared to the two GP models, the stark difference in superior predictions on the CHB source provided by LVGP can be also observed through the parity plots in Figure \ref{fig:beam_summary}c,d\&e.

\begin{table*}[h!]
\small
\centering
\caption{Prediction accuracy on cantilever beam sources using GP, LVGP, and GP-FSW models\label{tab:beam_pred}}
\renewcommand{\arraystretch}{1.3}
\begin{tabular}{ccccccccc}
\toprule
Model Type & Training NRMSE (All sources)  & Testing NRMSE (All sources) & Testing NRMSE (HCB) \\
\toprule
GP & $0.016$ & $0.033$  &  $0.056$ \\
\\
LVGP & $0.002$  & $0.019$  &  $0.031$\\
\\
GP-HCB & --&--& $0.065$\\
\toprule
\end{tabular}
\end{table*}

In addition to offering better predictions, the LVGP approach comes with the advantage of providing interpretability through latent space analysis. Figure \ref{fig:beam_summary}f shows the latent space and the distances between the sources to the CHB source. The distances between latent variables can help with possibly explaining the cause-effect relationship between sources. It is observed that the two sources that has the same underlying rectangular domain shape, RB and RHB, are located very close to each other whereas the HCB is located at a much further place. This is also observed through the dissimilarity metric ($D$) with HCB taken as the reference source for distance calculations. Even though the inputs are mapped onto the same domain, the LVGP approach can capture the underlying design similarities and differences between the sources through the latent space and therefore provide better predictions compared to existing approaches. Finally, although the underlying relationships between sources were known a priori for this case, this unique advantage provided by LVGP can be significantly beneficial to understand and interpret the sources of interests further for the cases where there is no prior information about the sources available.

\subsection{Case Study 2: Ellipsoidal Void Modeling with Varying Complexity and Fidelity Levels} \label{void}

\begin{figure}[h!]
      \centering
		\includegraphics[width=\linewidth]{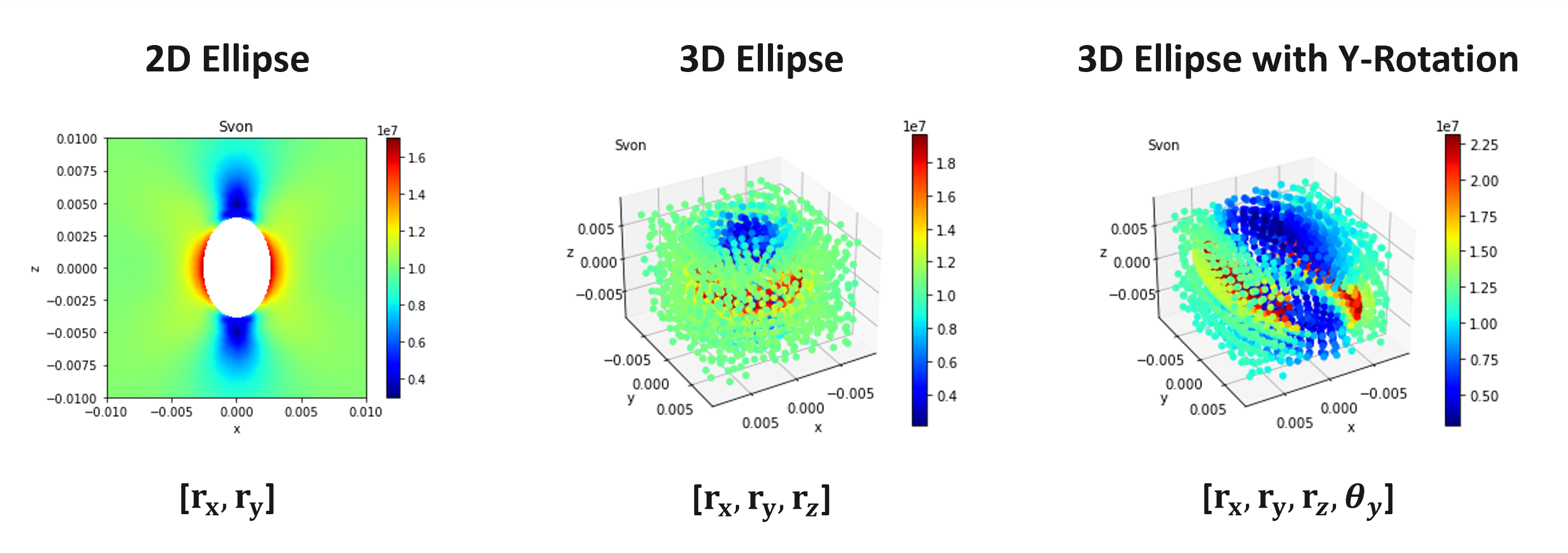}
		\caption{Ellipsoidal void sources with varying design complexity (2D \& 3D), and fidelity (elastic \& plastic)}
		\label{fig:Void}
\end{figure}

The second case study focuses on the design of 3D ellipsoidal voids, where the sources are set up not only to vary in design parametrization (as seen in the Cantilever Beam study) but also in the complexity (2D versus 3D) and fidelity of the structural simulation (elastic analysis versus plastic analysis). The schematics of the three sources are presented in Figure \ref{fig:Void}. For ease of notation, the 2D Ellipse, 3D Ellipse, and 3D Ellipse with Y-Rotation are named 2DE, 3DE, and 3DER, respectively. Similar to the previous case study, the sources vary in design complexity with additional features incorporated into the void parameterizations. Moreover, 3D differs from 3DE and 2DE in terms of simulation fidelity as well. Plastic structural analysis was used to evaluate designs for 3DER, while elastic structural analysis was employed for 2DE and 3DE. Consequently, the computational cost of data generation for 3DER is higher than the costs associated with 2DE and 3DE.

\subsubsection{Data Description}
The details of the dataset used for this case study are shown in Table \ref{tab:void_data}. The first source of data for the proposed data fusion framework is identified as the simplest and most computationally economical option for querying data. In this context, that would be 2DE (2D Ellipse) design, parameterized through $r_x$ and $r_z$, where $r_x$ and $r_z$ represent the axis lengths across the X and Z axes, respectively (see Figure \ref{fig:Void}). For 2DE, a total of 273 training data points and 117 testing data points were generated, and the structural analysis associated with this design is an elastic analysis. 3DE (3D Ellipse) is parameterized by $r_x$, $r_y$, and $r_z$, where $r_x$, $r_y$, and $r_z$ are the axis lengths across the X, Y, and Z axes, respectively. The structural analysis conducted for this source also remains as an elastic analysis. For 3DE, a total of 182 training points and 77 testing points were generated. 3DER (3D Ellipse with Y-Rotation) represents the final source of data, parameterized by $r_x$, $r_y$, $r_z$, and $\theta_y$, where $r_x$, $r_y$, and $r_z$ are the axis lengths across the X, Y, and Z axes, respectively, and $\theta_y$ is the angle of rotation of the void across the Y axis. The structural analysis associated with this third source of data is a plastic analysis under linearly elastic and plastic conditions. A total of 10 training points and 40 testing points were generated for 3DER. Finally, the output of interest for all design sources is the maximum Von Mises stress ($\sigma_{VM}$), spatially determined.

\begin{table*}[h]
\small
\centering
\caption{Ellipsoidal void data description\label{tab:void_data}}
\renewcommand{\arraystretch}{1.3}
\begin{tabular}{>{\centering\arraybackslash}p{3cm} >{\centering\arraybackslash}p{3cm} >{\centering\arraybackslash}p{2cm} >{\centering\arraybackslash}p{3cm} >{\centering\arraybackslash}p{3cm}}
\toprule
Source Type & Input Variables &Output Variable& Number of Training Samples  & Number of Testing Samples \\
\toprule
2D Ellipse (2DE) & $r_x (m),r_z (m)$ & & $273$  &  $117$\\
\\
3D Ellipse (3DE) & $r_x (m)$$,r_y (m)$$,r_z (m)$ &Maximum Von Mises Stress ($\sigma_{VM}$) &$182$ & $77$\\
\\
3D Ellipse with Y-Rotation (3DER) & $r_x (m),r_y (m),r_z (m),$ $\theta_y (degree)$ & &$10$& $40$\\
\toprule
\end{tabular}
\end{table*}

\subsubsection{Results}
\begin{figure}[h]
      \centering
	   \begin{subfigure}{0.45\linewidth}
		\includegraphics[width=\linewidth]{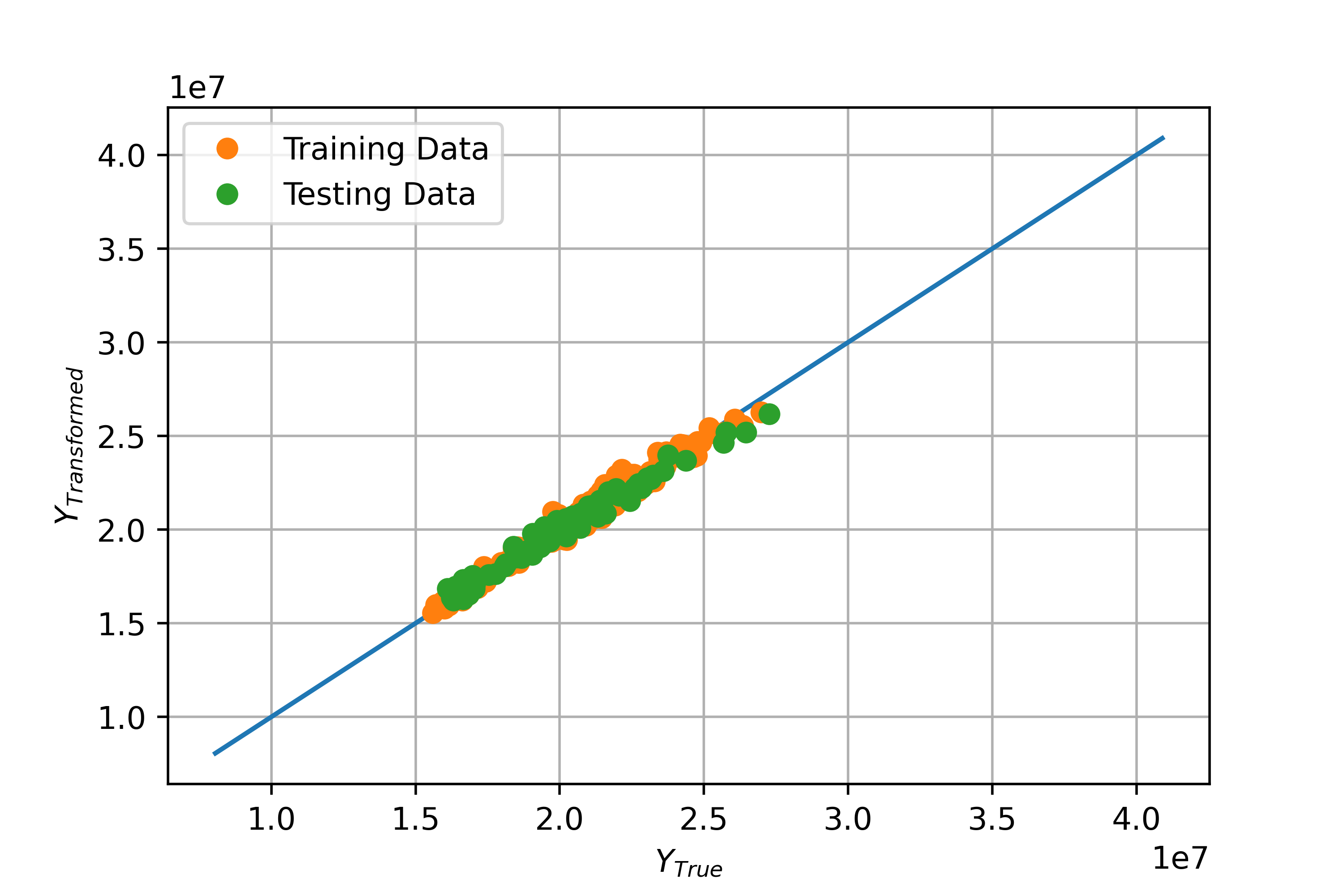}
		\caption{}
		\label{fig:sub1}
	   \end{subfigure}
	   \begin{subfigure}{0.45\linewidth}
		\includegraphics[width=\linewidth]{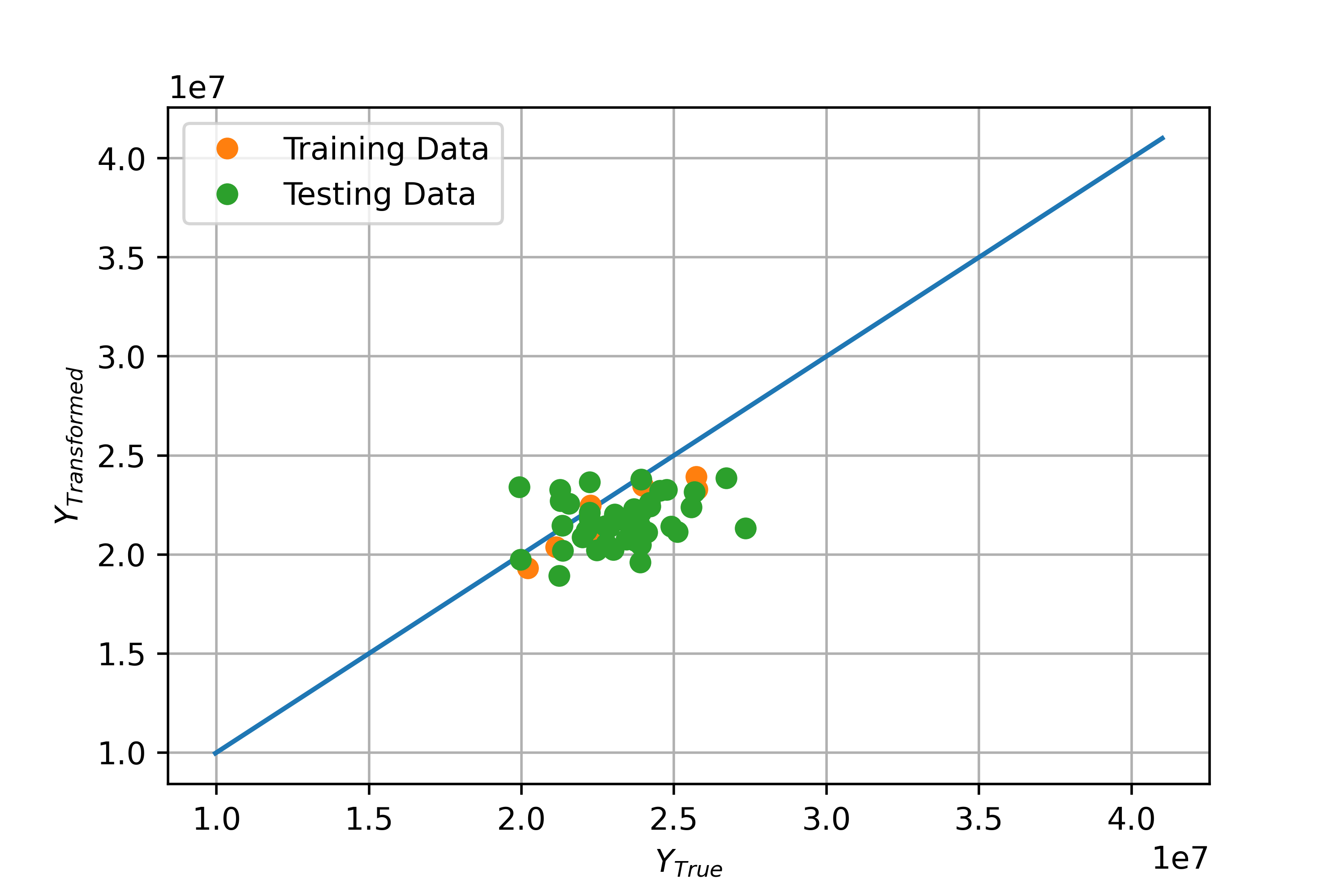}
		\caption{}
		\label{fig:sub2}
	    \end{subfigure}
	\caption{Predictions based on mapped (a) 3D Ellipse and (b) 3D Ellipse with Y-rotation sources using the reference GP model built on 2D Ellipse source}
	\label{fig:void_transform}
\end{figure}

\paragraph{Heterogeneous Mapping}
As highlighted in the methodology section, the initial step of the proposed heterogeneous multi-source framework is to identify the transformations between the sources and the reference source (2DE). A GP model is first trained on the data from 2DE, and this trained model is then utilized to optimize the transformation matrix $A$ and vector $b$ for the input parameters of 3DE and 3DER relative to 2DE. Equation \ref{eqn:ellip_void_t1} demonstrates the optimized transformation of 3DE onto 2DE. The first observation that can be made is that $r_z$ is preserved across both sources, as indicated by the entry of 0.99 in the $A$ matrix and 0.00 in $b$. The equivalence of maximum Von Mises stress in 3DE, derived from $r_x$ and $r_y$, is captured through $r_x$ in 2DE as a direct sum of $r_x$ and $r_y$, with a shift of 0.49. Furthermore, the prediction performance of the transformation from 3DE using the 2DE GP model is presented in Figure \ref{fig:void_transform}a. It can be observed that the transformation provides a very good fit for the training data and generalizes well to the testing data. The same process is repeated for the transformation between 3DER and 2DE, with the optimized transformation demonstrated through Equation \ref{eqn:ellip_void_t2}. It is initially observed that the $r_z$ equivalence of 3DER in 2DE does not have a shift parameter. The transformed $r_z$ is primarily determined by $r_z$ (with a scaled weight of 0.64) and mildly by $\theta_y$ (with a scaled weight of 0.09). The transformed $r_x$ is influenced by $r_x$, $r_y$, and $r_z$ from 3DER, with no influence from $\theta_y$, and includes a significant shift driven by a value of 2.00. The predictive performance of the transformation from 3DER using the 2DE GP model is shown in Figure \ref{fig:void_transform}b. There is noticeable scatter across the 45-degree line for the 2DE equivalence of 3DER. The authors believe that this scatter is primarily driven by the difference in the fidelities of the simulation where 2DE uses elastic analysis, while 3DER employs plastic analysis.

\begin{equation}\label{eqn:ellip_void_t1}
\begin{bmatrix} r_x \\ r_z \end{bmatrix}_{Norm} = \begin{bmatrix} 0.98 & 0.98 & 0.00 \\ 0.00 & 0.00 & 0.99 \end{bmatrix}  \begin{bmatrix} r_x \\ r_y \\ r_z \end{bmatrix}_{Norm} + \begin{bmatrix} 0.49 \\ 0.00 \end{bmatrix} \\
\end{equation}

\begin{equation}\label{eqn:ellip_void_t2}
\begin{bmatrix} r_x \\ r_z \end{bmatrix}_{Norm} = \begin{bmatrix} 0.80 & 0.47 & 1.13 & 0.00 \\ 0.00 & 0.00 & 0.64 & 0.09 \end{bmatrix}  \begin{bmatrix} r_x \\ r_y \\ r_z \\ \theta_y \end{bmatrix}_{Norm} + \begin{bmatrix} 2.00 \\ 0.00 \end{bmatrix} \\
\end{equation}

\begin{figure}[h]
      \centering
	   \begin{subfigure}{0.45\linewidth}
		\includegraphics[width=\linewidth]{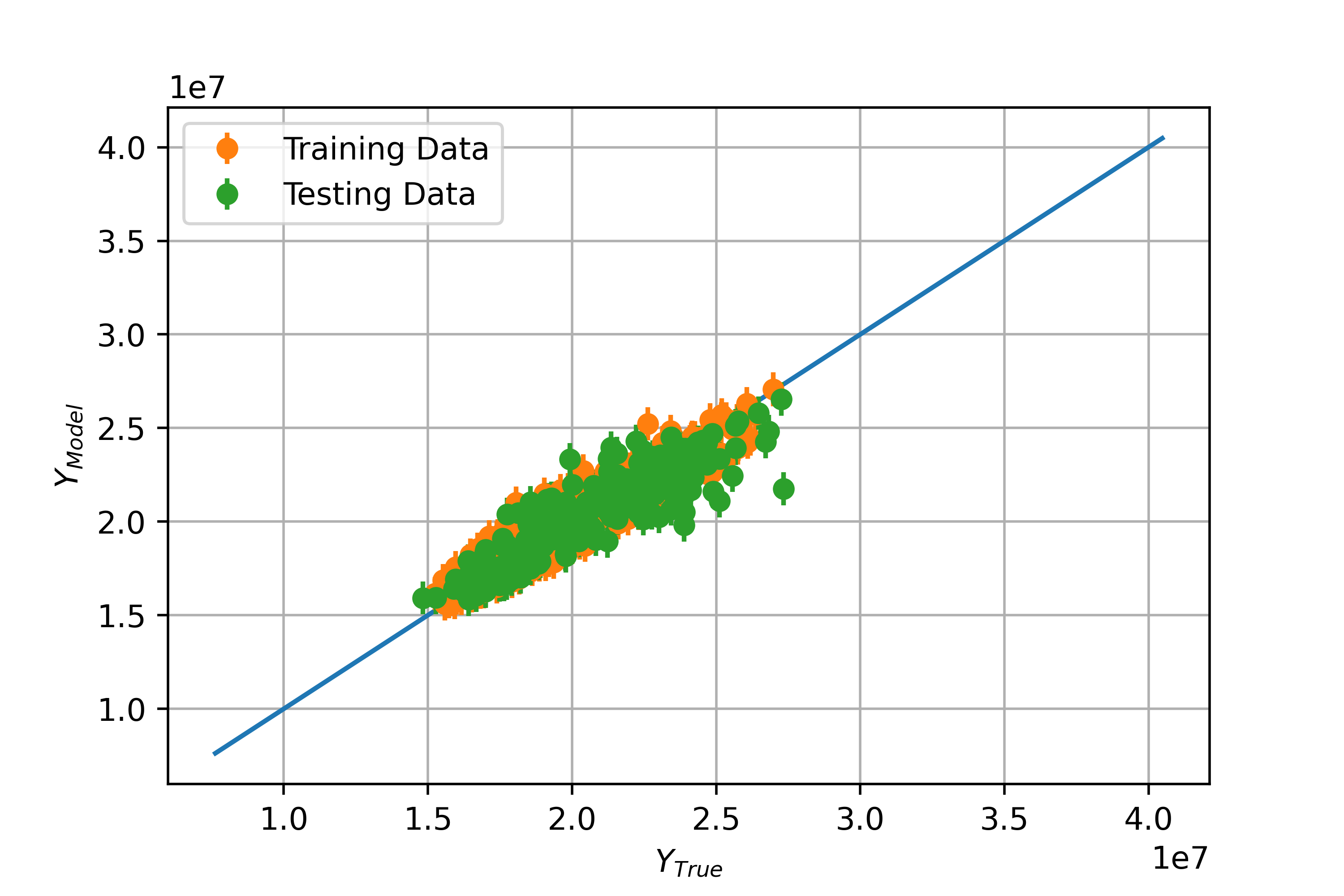}
		\caption{}
		\label{fig:subfig2}
	    \end{subfigure}
	     \begin{subfigure}{0.45\linewidth}
		 \includegraphics[width=\linewidth]{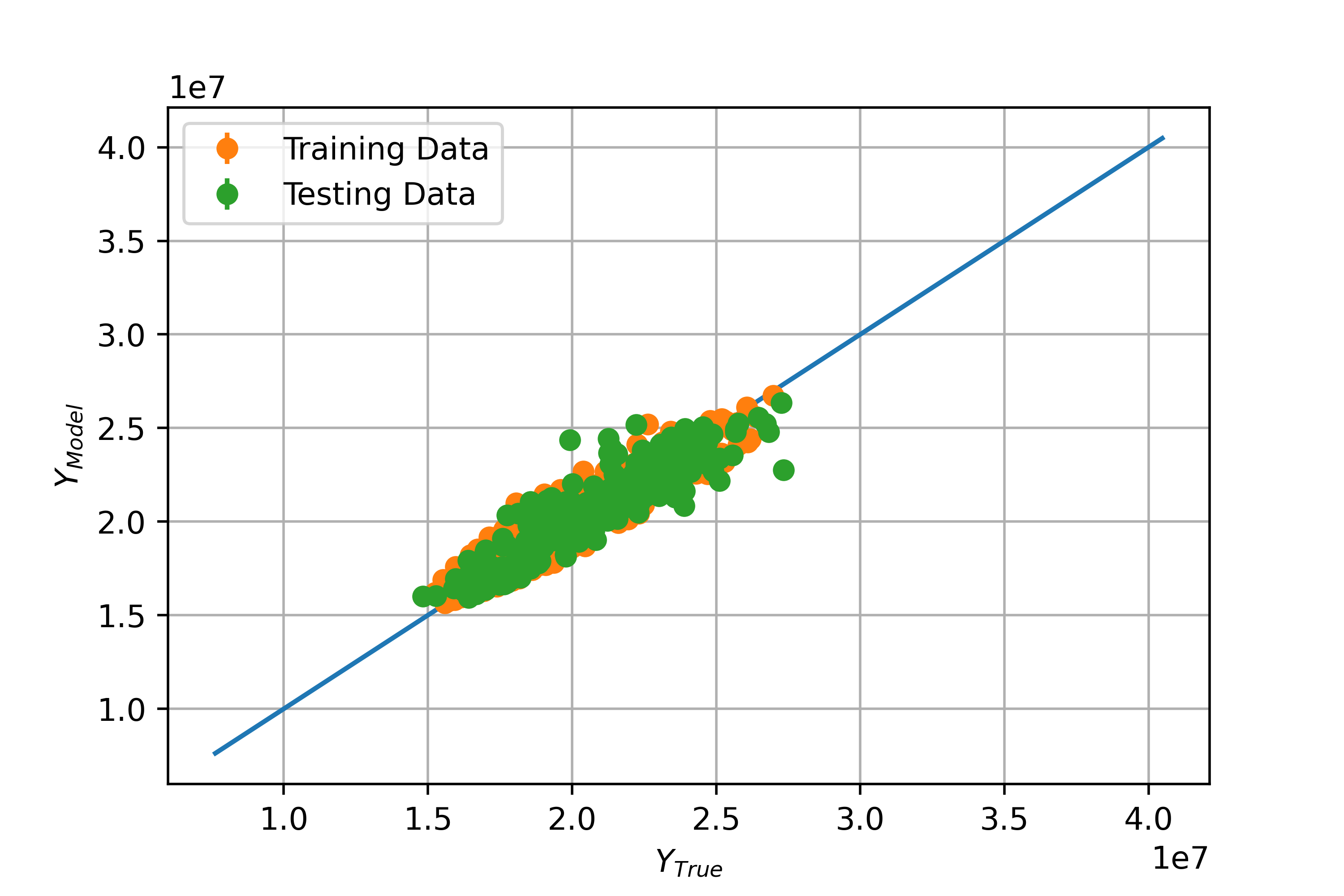}
		 \caption{}
		 \label{fig:subfig3}
	      \end{subfigure}
	      \vfill
	     \begin{subfigure}{0.33\linewidth}
		\includegraphics[width=\linewidth]{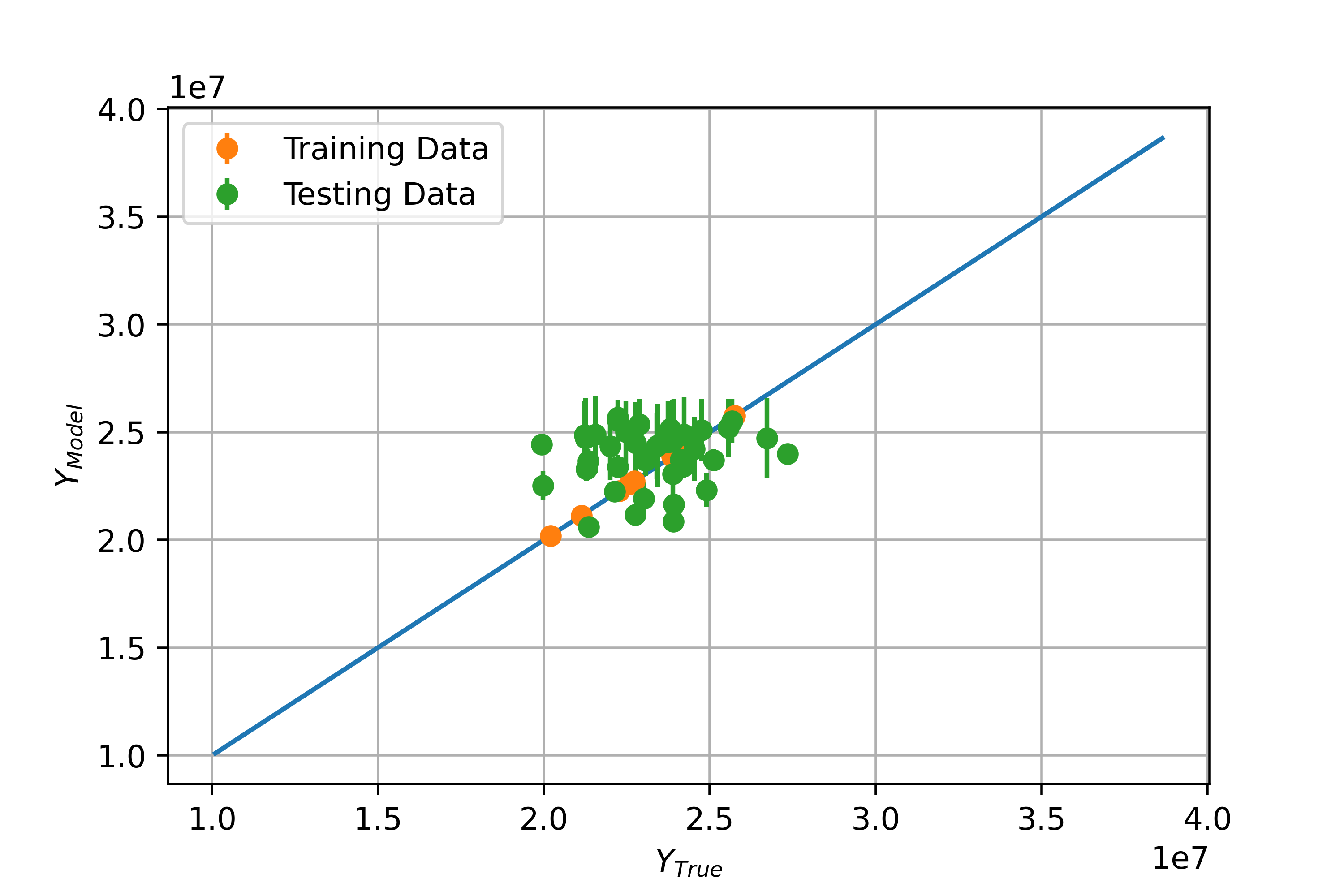}
		\caption{}
		\label{fig:subfig1}
	   \end{subfigure}
	   	     \begin{subfigure}{0.33\linewidth}
		\includegraphics[width=\linewidth]{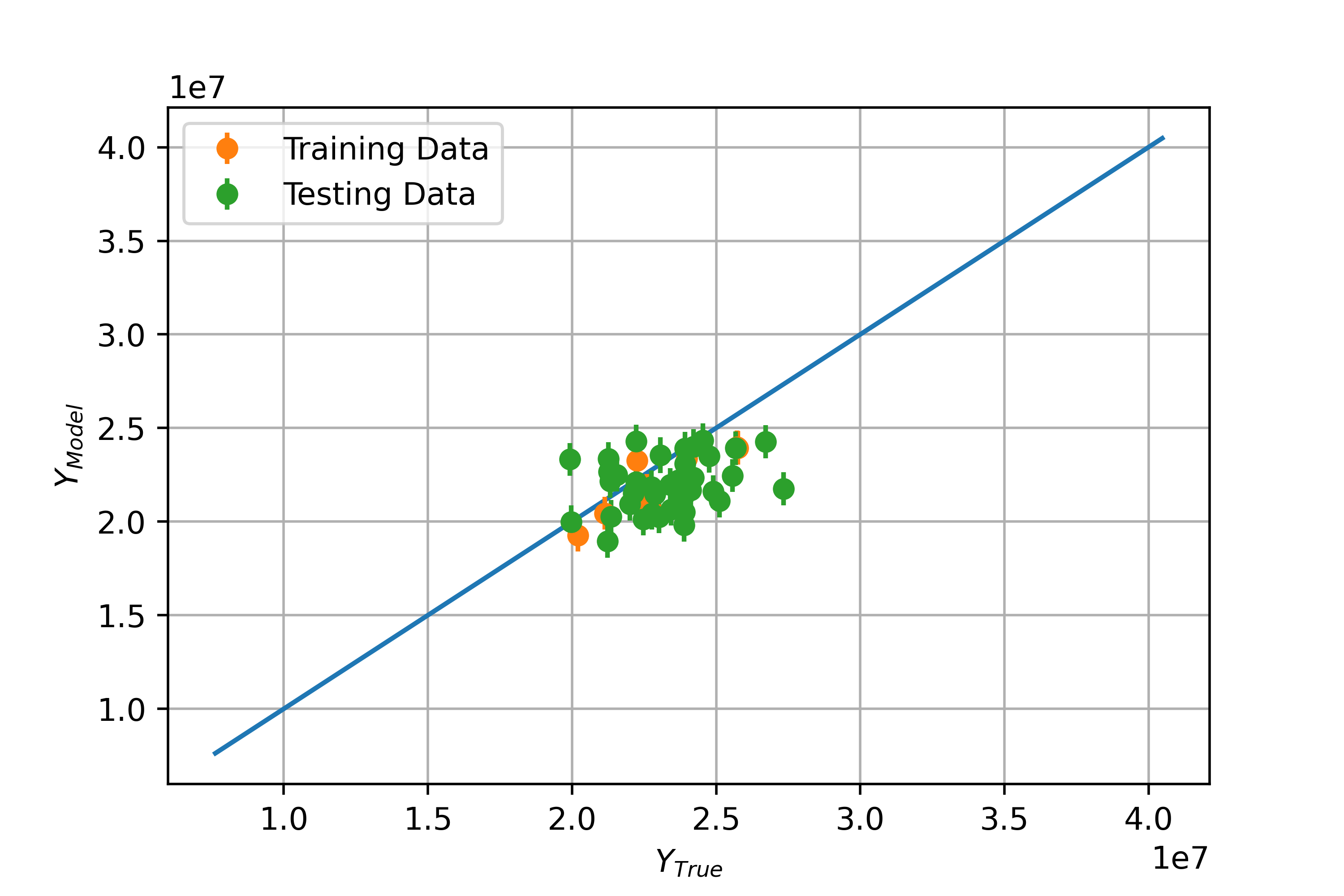}
		\caption{}
		\label{fig:subfig1}
	   \end{subfigure}
	   	     \begin{subfigure}{0.33\linewidth}
		\includegraphics[width=\linewidth]{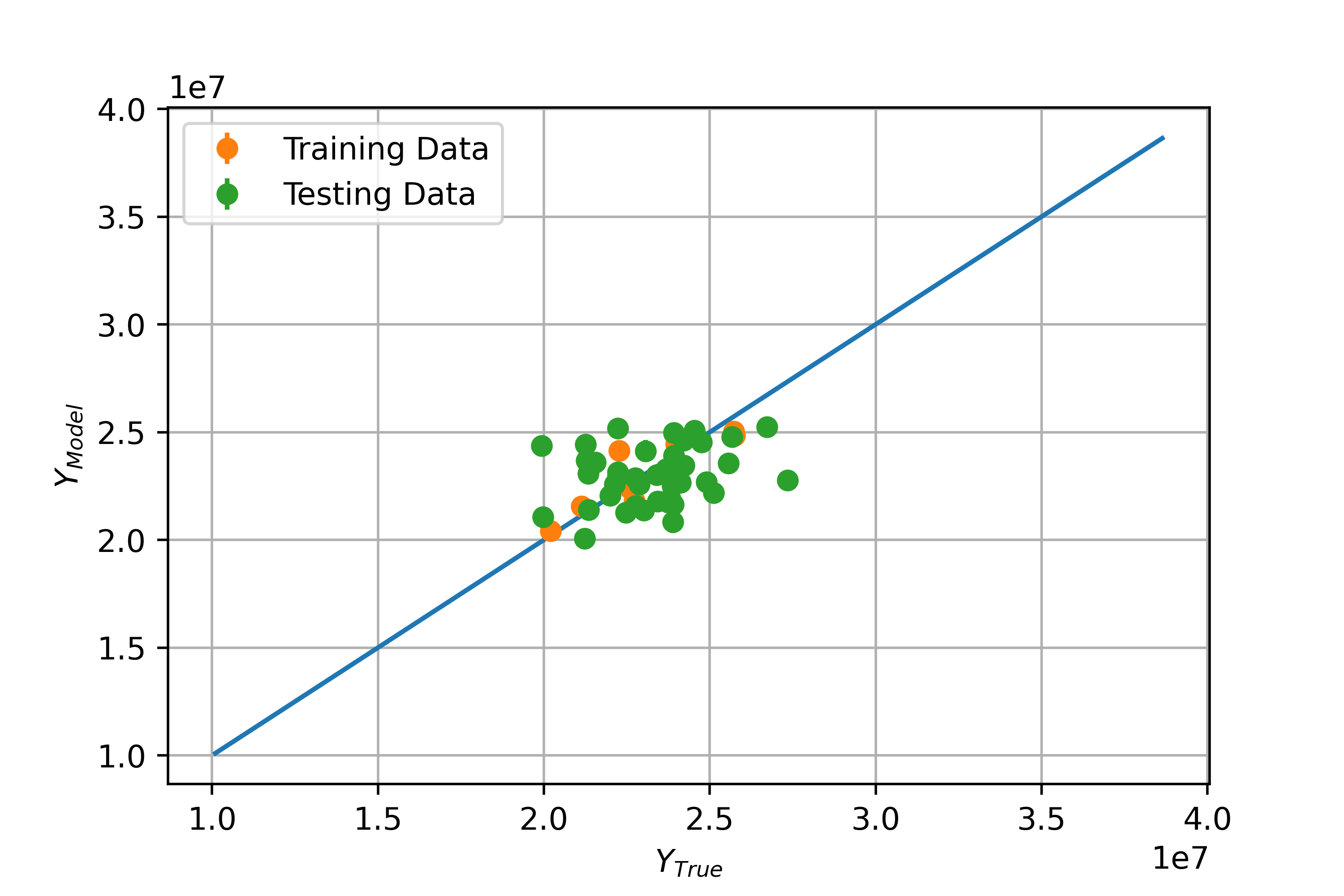}
		\caption{}
		\label{fig:subfig1}
	   \end{subfigure}
	   
	   \vfill
	       \begin{subfigure}{0.45\linewidth}
		  \includegraphics[width=\linewidth]{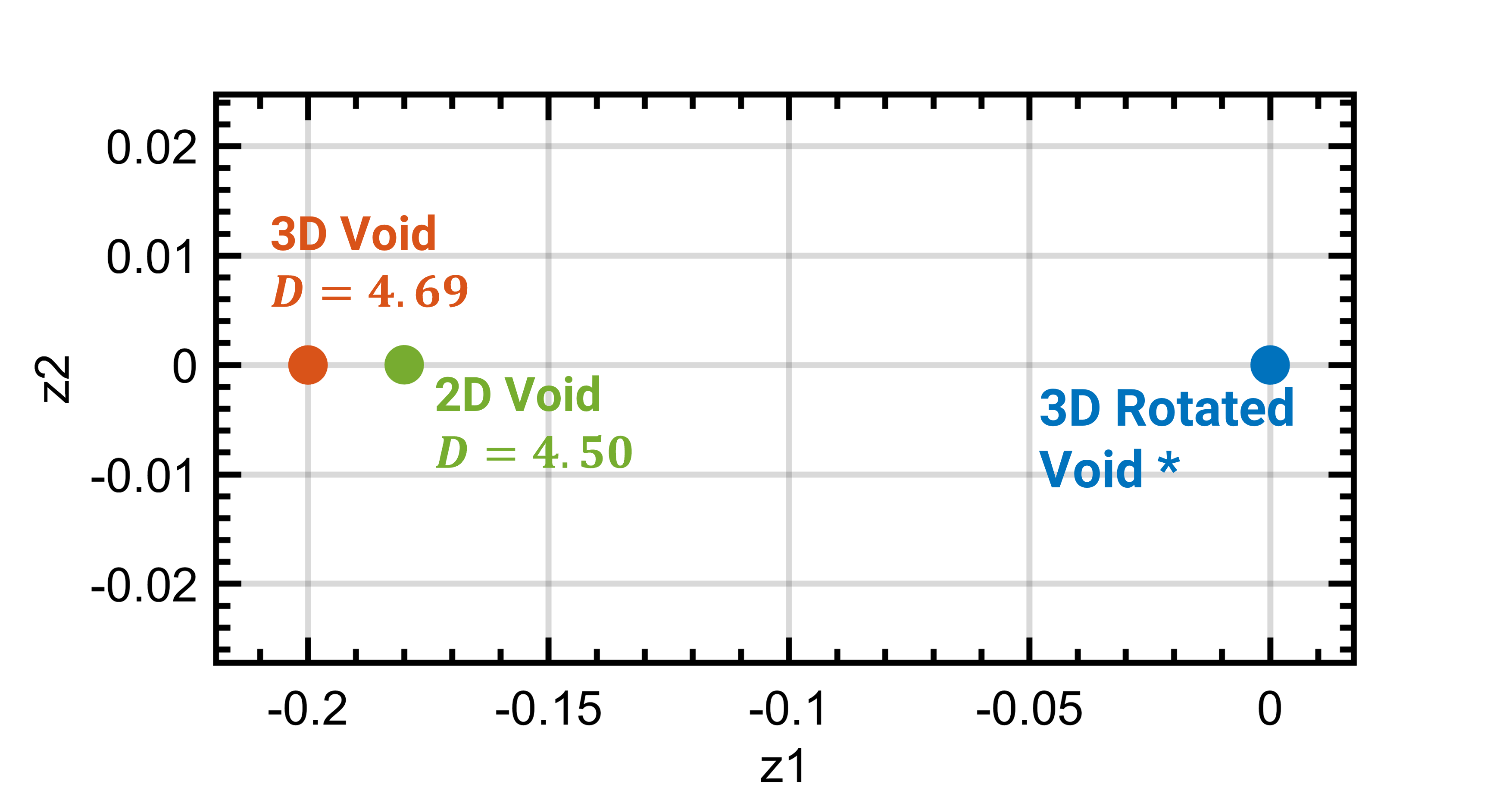}
		  \caption{}
		  \label{fig:subfig4}
	       \end{subfigure}
	\caption{The results of the ellipsoidal void design study. Predictions on (a) all sources using heterogeneous multi-source GP, (b)  sources using heterogeneous multi-source LVGP, (c) hollow circular  beam using single-source GP built on original input space, (d) hollow circular beam using heterogeneous multi-source GP, (e) hollow circular beam using heterogeneous multi-source LVGP. (f) The latent space obtained by the LVGP model}
	\label{fig:subFigures_void}
\end{figure}

\paragraph{Multi-Source Modeling Comparison}
To highlight, compare, and analyze the performance of the proposed approach, similar to the previous case study, three different modeling approaches are proposed. Initially, multi-source models, GP and LVGP, are compared for generalizability (predictions on all three sources) and for the source with the least amount of data (3DER). Through the transformation presented as the first step of the framework, both models are trained and tested on the same data, with the parameter space defined by 2DE. It is important to highlight again that with the introduction of LVGP, each transformed and the reference data is treated as a separate source, whereas this distinction is not made by the GP model. The prediction results on the training and testing data are presented in Table \ref{tab:void_pred} and Figure \ref{fig:subFigures_void}. From the NRMSE metric, it can be seen that LVGP outperforms the GP model in both the training set (by $2.7\%$) and testing set (by $10.4\%$) for predictions across all three sources. Furthermore, LVGP model provides lower prediction uncertainties compared to the GP model (Figure \ref{fig:subFigures_void}). This result demonstrates that LVGP is a more generalizable model for the case study of sources with different fidelities and complexities. Moreover, examining the prediction results on 3DER only, a significant improvement (of $19.3\%$) in the prediction metric is observed. This improvement is achieved by differentiating between sources and transferring knowledge from various sources to make better source-specific predictions. Next, a GP model based on the original input parameters of 3DER is built and compared with the multi-source LVGP on the 3DER testing data only. This single-source GP model is referred to as GP-3DER in Table \ref{tab:void_pred}. Once again, it is observed that the LVGP approach provides improvements in prediction metrics (of $10.8\%$), even when GP-3DER was trained over the original 3DER data. The parity plots in Figure \ref{fig:subFigures_void}c,d\&e clearly shows the notable improvement in predictions for the 3DER source provided by LVGP compared to the two alternative GP models. The comparison and reduction in errors also highlight the application of the proposed framework for highly data-scare scenarios. For our current case study, only 10 training datapoints is made available. By addition of other relatively cheaper sources of information, a combined generalizable model can be developed through the proposed framework.

Even though all the sources' inputs were brought to the same parameter space through IMC, the underlying relationship between the sources and the output might still differ from each other. GP modeling lacks an inherent mechanism to handle this variability, whereas LVGP addresses it through the learned latent variables. This is further evidenced in the latent space of LVGP, as shown in Figure \ref{fig:subFigures_void}f. Taking the 3DER source as a reference for dissimilarity metric ($D$) calculations, the 2DE and 3DE sources exhibit noticeable dissimilarities of $4.89$ and $4.49$, respectively, from the reference source. The almost identical difference is primarily driven by the discrepancy between plastic and elastic analyses. The relative difference of $0.20$ between the two elastic sources is driven by the differences between 2D and 3D modeling. Therefore, this inherent interpretability of LVGP bridges the modeling gap between the sources and provides an accurate modeling advantage. Finally, for other potential problems that may or may not vary in fidelity, the differences created through design variables might be too significant for IMC or any other mapping technique to fully capture. There might also be situations where input mapping itself might bridge the gap between the sources. As a result, all these factors can be understood from the latent space and the dissimilarity metric provided by the LVGP approach.

\begin{table*}[h]
\small
\centering
\caption{Prediction accuracy on ellipsoidal void sources using GP, LVGP, and GP-3DER models \label{tab:void_pred}}
\renewcommand{\arraystretch}{1.3}
\begin{tabular}{ccccccccc}
\toprule
Model Type & Training NRMSE (All sources)  & Testing NRMSE (All sources) & Testing NRMSE (3DER) \\
\toprule
GP & $0.074$ & $0.105$  &  $0.409$ \\
\\
LVGP & $0.072$  & $0.094$  &  $0.330$\\
\\
GP-3DER & --&--& $0.370$\\
\toprule
\end{tabular}
\end{table*}

\vspace{0.8cm}
\subsection{Case Study 3: Manufacturing Process Fusion of Ti6Al4V Alloys} \label{titanium}
Titanium alloy (Ti6Al4V) has been at the forefront of aerospace applications due to its superior corrosion resistance, strength, and toughness. The manufacturing of this alloy has evolved beyond the conventional casting process, extending to additive and solid-state manufacturing processes such as Electron Beam Melting (EBM), Laser Powder Bed Fusion (LPBF), and Friction Stir Welding (FSW). In the third and final case study, a multi-source data fusion model across different manufacturing modalities, for the same alloy is demonstrated. Specifically, three thermally driven manufacturing processes, EBM, LBPF, and FSW are selected as the data sources. A special characteristics of this study is that there is a complete non-overlap in the input parameter space of the data sources, meaning that there is no common parameter across sources. It is important to note that the complexity and cost of each source are open to interpretation based on the modeling scenario. The fundamental question this work seeks to answer is whether learning across different manufacturing modalities is possible for the same alloy through heterogeneous multi-source data fusion.



\subsubsection{Data Description}
The manufacturing data for the three domains are obtained from prior published works \cite{luo2022effect,luo2023dataset,ran2020microstructure,fall2017effect} and the details of the dataset are provided in Table \ref{tab:titanium_data} . Based on data availability, FSW is set to be the Source 3. LPBF is selected as Source 1 and the reference source for mapping purposes. EBM is referenced as Source 2. Finally, the output to be modeled for all three modes of manufacturing is the yield strength ($sigma_y$) of the Ti6Al4V alloy.



\begin{table*}[h]
\small
\centering
\caption{Ti6Al4V data description\label{tab:titanium_data}}
\renewcommand{\arraystretch}{1.3}
\begin{tabular}{>{\centering\arraybackslash}p{3.2cm} >{\centering\arraybackslash}p{3.1cm} >{\centering\arraybackslash}p{2cm} >{\centering\arraybackslash}p{3cm} >{\centering\arraybackslash}p{3cm}}
\toprule
Source Type & Input Variables & Output Variable & Number of Training Samples & Number of Testing Samples \\
\toprule
Laser Powder Bed Fusion (LBPF) \cite{luo2022effect,luo2023dataset} & \textit{Laser Power (W)}, \hspace{20mm}\textit{Laser Speed (mm/s)} & & $29$  &  $13$ \\
\\
Electron Beam Melting (EBM) \cite{ran2020microstructure}  & \textit{Focus Offset (mA)}, \hspace{20mm}\textit{Line Offset (mm)},\hspace{20mm} \textit{Speed Factor} & Yield Strength ($\sigma_y$, MPa) & $10$ & $5$ \\
\\
Friction Stir Welding (FSW) \cite{fall2017effect} & \textit{Rotational Speed (RPM)}, \textit{Travel Speed (mm/min)} & & $3$ & $6$ \\
\toprule
\end{tabular}
\end{table*}


\subsubsection{Results}
\begin{figure}[ht]
      \centering
	   \begin{subfigure}{0.45\linewidth}
		\includegraphics[width=\linewidth]{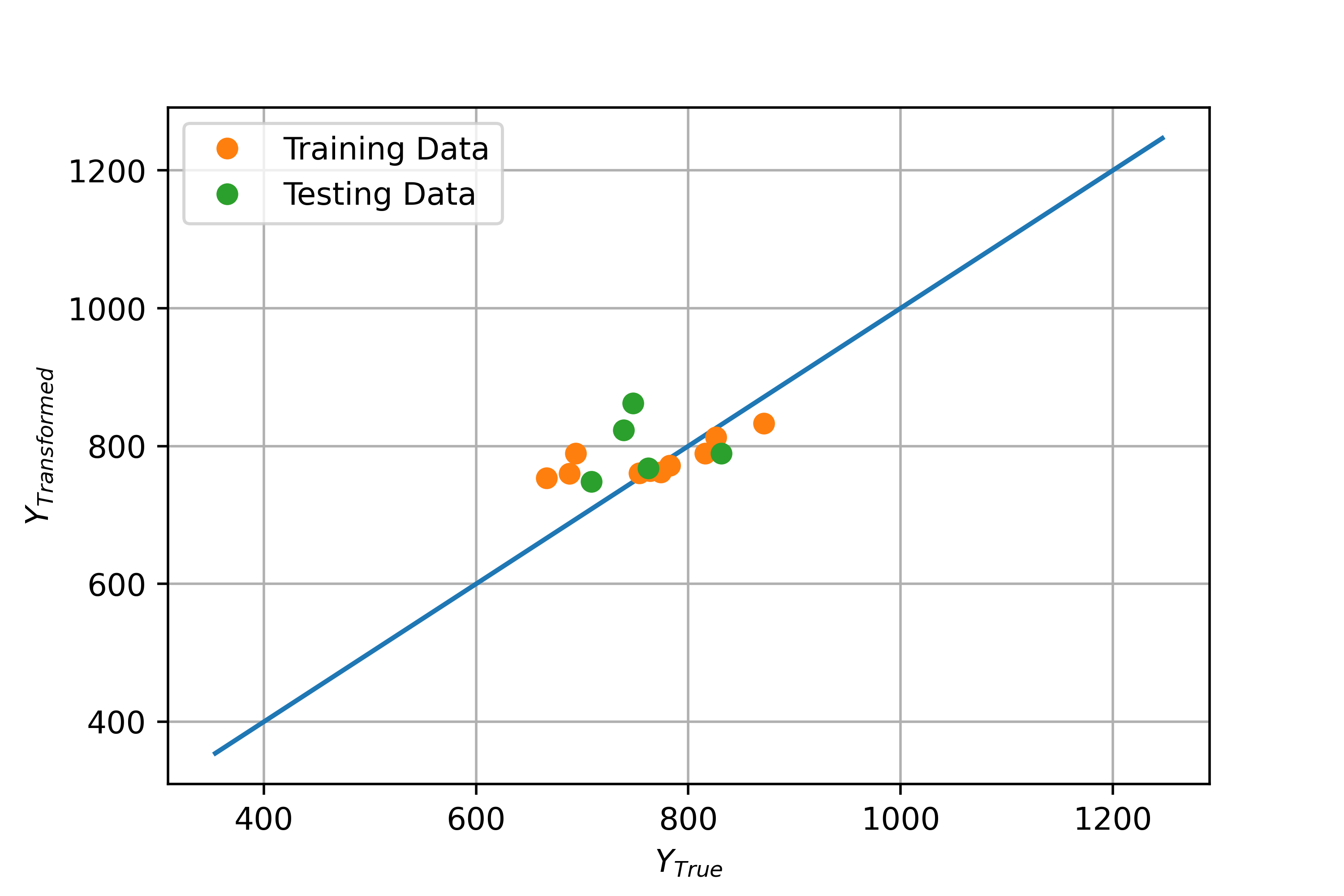}
		\caption{}
		\label{fig:subfig1}
	   \end{subfigure}
	   \begin{subfigure}{0.45\linewidth}
		\includegraphics[width=\linewidth]{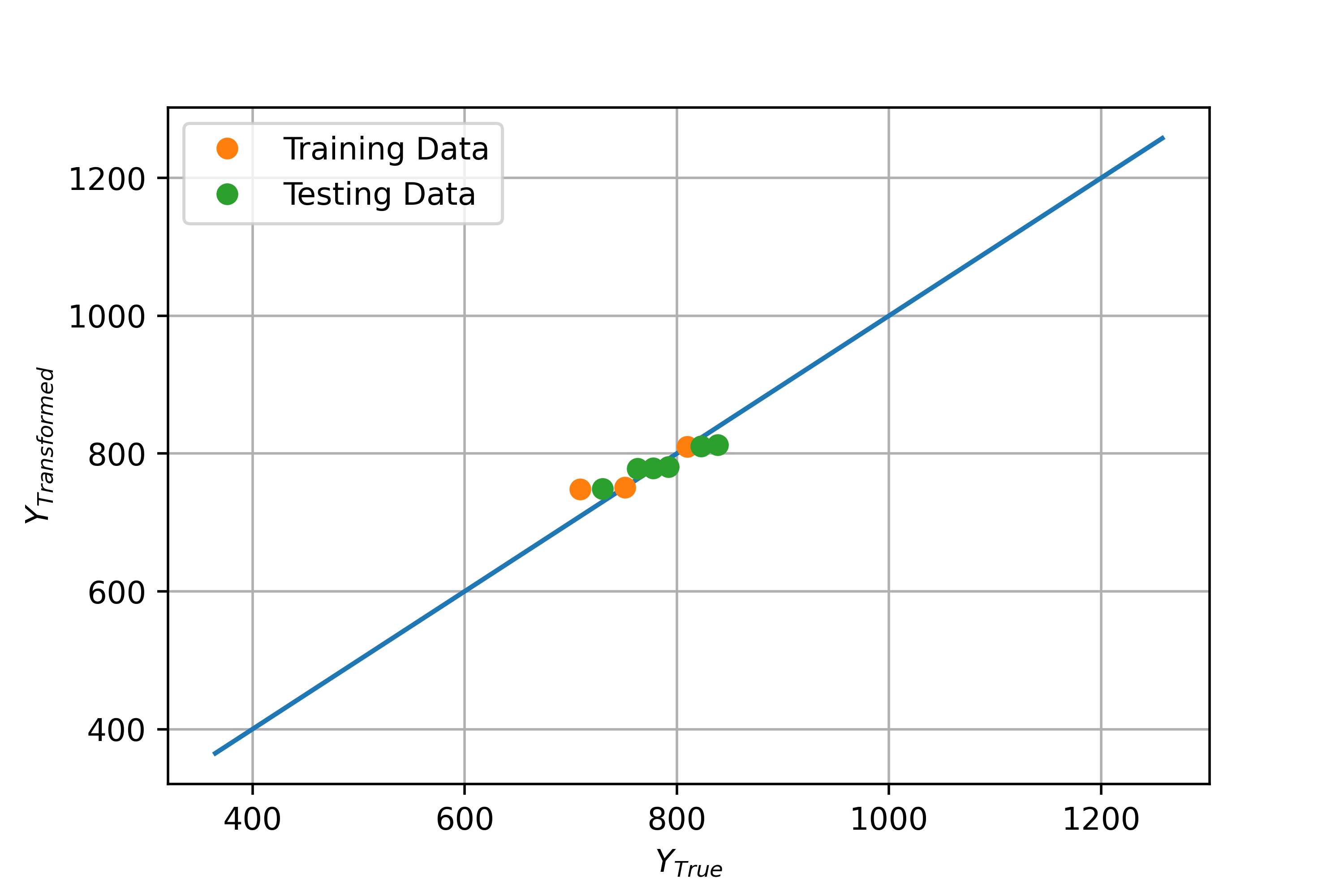}
		\caption{}
		\label{fig:subfig2}
	    \end{subfigure}
	\caption{Predictions based on mapped (a) electron beam welding and (b) friction stir welding sources using the reference
GP model built on laser bed powder fusion source}
	\label{fig:ti_transform}
\end{figure}

\paragraph{Heterogeneous Mapping} 
The initial step of the proposed framework involves identifying the mapping between EBM and FSW to the LPBF input domain. A standalone GP model is trained on the LPBF data, which is then utilized in the optimization framework to construct the $A$ and $b$ matrices. The transformations obtained for EBM and FSW to LPBF are detailed through Equations \ref{eqn:ti6al4v_t1} and \ref{eqn:ti6al4v_t2}. The transformation of EBM variables to the LPBF domain is primarily influenced by the \textit{Focus Offset} and \textit{Line Offset} variables, with a minimal contribution from the \textit{Speed Factor}. Intuitively, this transformation aligns with the understanding of the amount of heat input into the system, which can significantly influence the yield strength of the material. Specifically, this heat amount is primarily driven by \textit{Laser Power} for LPBF and a combination of \textit{Focus Offset} and \textit{Line Offset} for EBM. Both outputs are shifted by noticeable factors. The \textit{Laser Power} equivalent for FSW is primarily derived from \textit{Travel Speed}. Furthermore, the Laser Speed equivalent for FSW is influenced by both the \textit{Rotational Speed} and \textit{Travel Speed} of the friction stir welding tool. A significant shift exists for this transformation as well. Finally, the parity plot displaying the predictive results using the reference LPBF GP model with the transformed input space of the two other sources is shown in Figure \ref{fig:ti_transform}. The transformed EBM predictions are acceptable, with higher point deviations observed in data points with yield strength less than 800 MPa approximately (Figure \ref{fig:ti_transform}a). For the FSW transformation predictions, the fit also demonstrates acceptable accuracy (Figure \ref{fig:ti_transform}b).

\begin{equation}\label{eqn:ti6al4v_t1}
\begin{bmatrix} Laser \ Power \\ Laser \ Speed \end{bmatrix}_{Norm} = \begin{bmatrix} 0.22 & 0.15 & 0.04 \\ 0.06 & 0.33 & -0.29 \end{bmatrix}  \begin{bmatrix} Focus \ Offset \\ Line \ Offset \\ Speed \ Factor \end{bmatrix}_{Norm} + \begin{bmatrix} -2.62 \\ -1.81 \end{bmatrix} \\
\end{equation}

\begin{equation}\label{eqn:ti6al4v_t2}
\begin{bmatrix} Laser \ Power \\ Laser \ Speed  \end{bmatrix}_{Norm} = \begin{bmatrix} 0.01 & 0.45 \\ -0.12 & -0.09 \end{bmatrix}  \begin{bmatrix} Rotational \ Speed \\ Travel \ Speed \end{bmatrix}_{Norm} + \begin{bmatrix} -3.15 \\ -2.15 \end{bmatrix} \\
\end{equation}

\paragraph{Multi-Source Modeling Comparison}
In line with the previous case studies, three modeling paradigms were evaluated using data from three distinct manufacturing process sources. When comparing the multi-source models, GP and LVGP, trained and tested on the original LPBF and transformed EBW and FSW data (as presented in Table \ref{tab:ti_pred}), it was observed that the GP model shows better performance on the training set for all three sources. On the other hand, the LVGP outperforms the GP on the testing set (by $14.4\%$). This significant improvement establishes the LVGP model as the preferred model for data fusion modeling once again. Finally, for predictions targeting the FSW source, where the amount of data available is significantly limited, the LVGP once again excels in terms of predictive capabilities compared to the GP model by learning the relationship between the available source information through the latent variables.

\begin{table*}[h!]
\small
\centering
\caption{Prediction accuracy on Ti6Al4V sources using GP, LVGP, and GP-FSW models \label{tab:ti_pred}}
\renewcommand{\arraystretch}{1.3}
\begin{tabular}{ccccccccc}
\toprule
Model Type & Training NRMSE (All sources)  & Testing NRMSE (All sources) & Testing NRMSE (FSW) \\
\toprule
GP & $0.057$ & $0.125$  &  $0.420$ \\
\\
LVGP & $0.55$  & $0.107$  &  $0.134$\\
\\
GP-FSW & --&--& $0.336$\\
\toprule
\end{tabular}
\end{table*}

\begin{figure}[h]
      \centering
	   \begin{subfigure}{0.45\linewidth}
		\includegraphics[width=\linewidth]{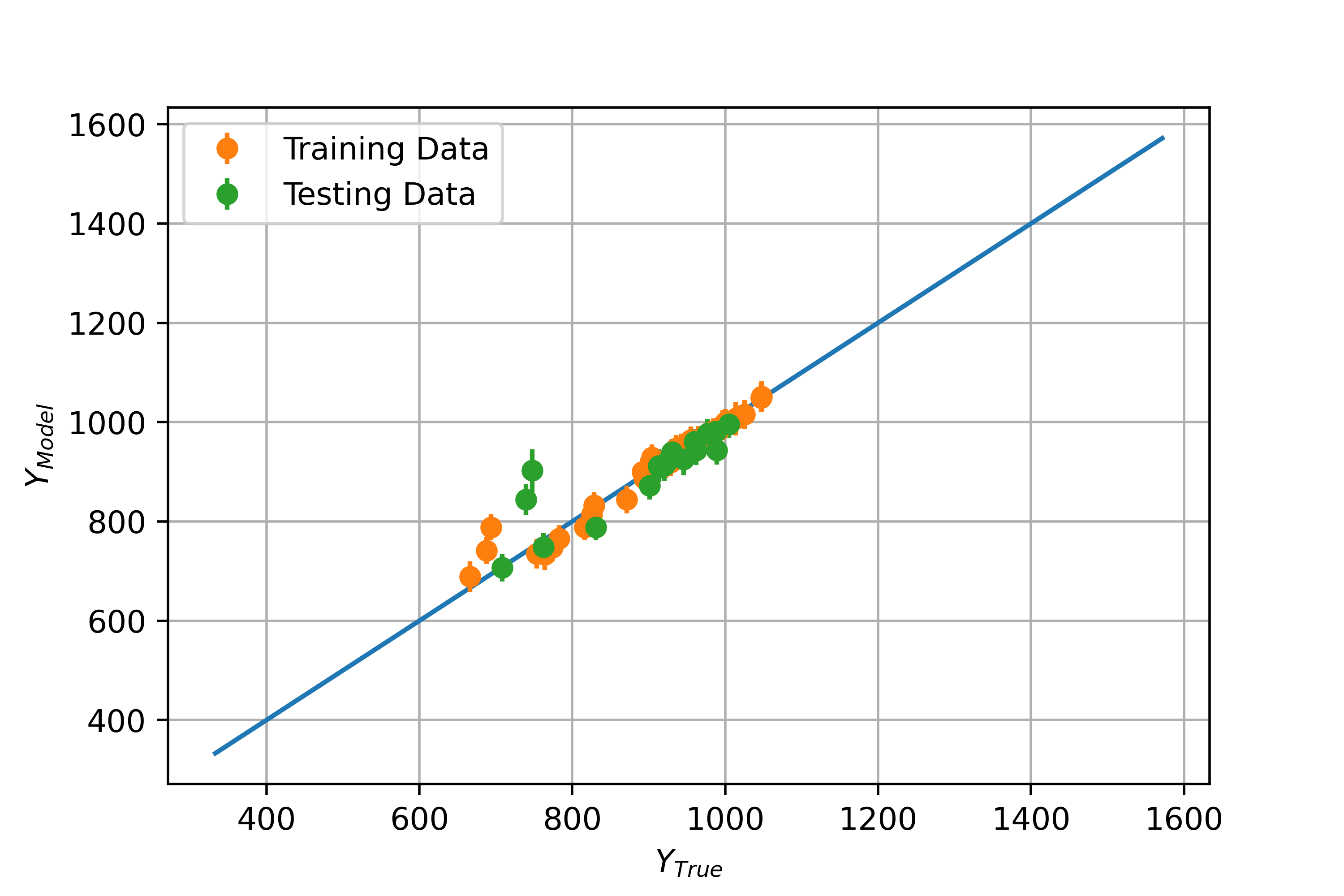}
		\caption{}
		\label{fig:subfig2}
	    \end{subfigure}
	    	  \begin{subfigure}{0.45\linewidth}
		 \includegraphics[width=\linewidth]{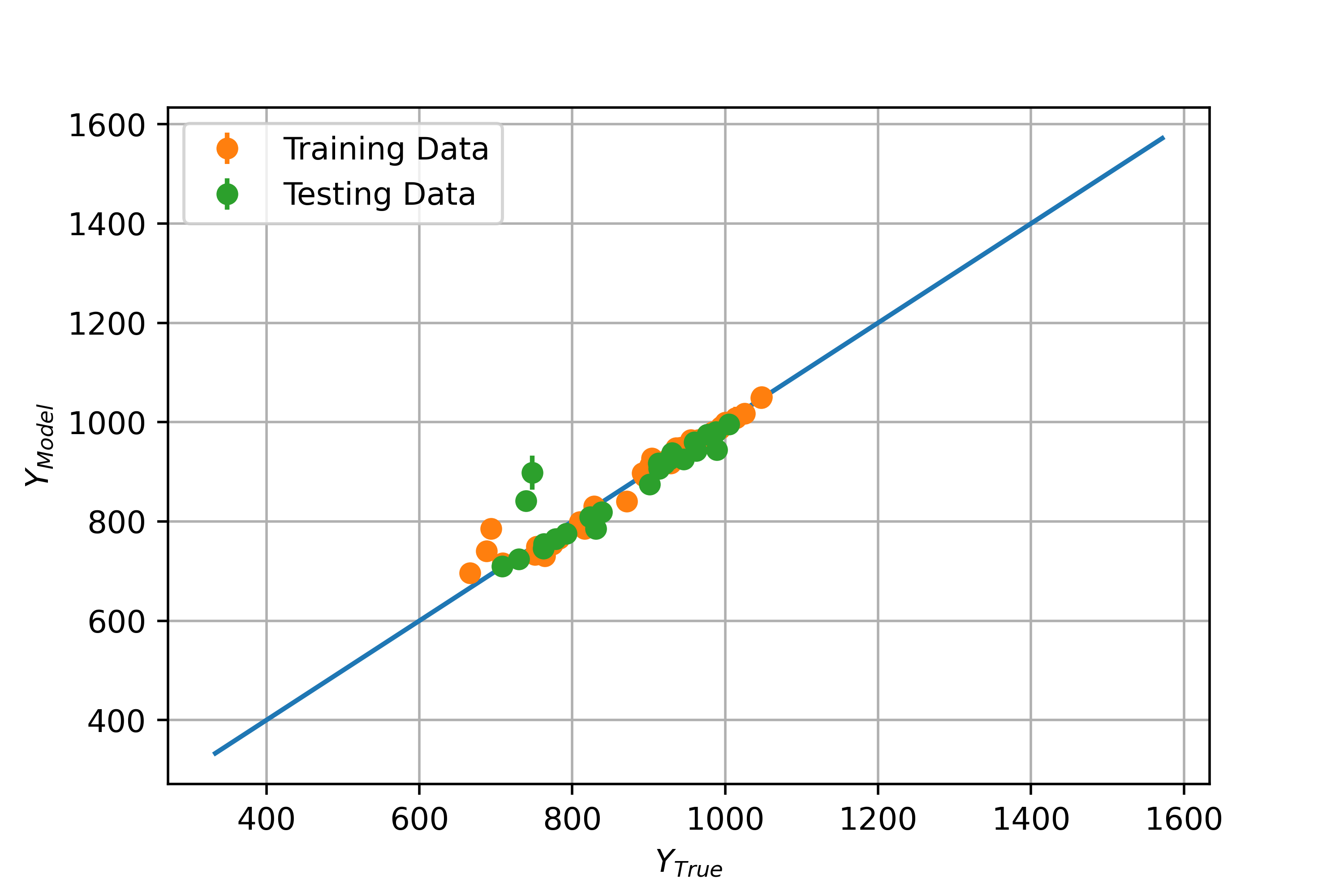}
		 \caption{}
		 \label{fig:subfig3}
	      \end{subfigure}
	\vfill
	   	       \begin{subfigure}{0.33\linewidth}
		\includegraphics[width=\linewidth]{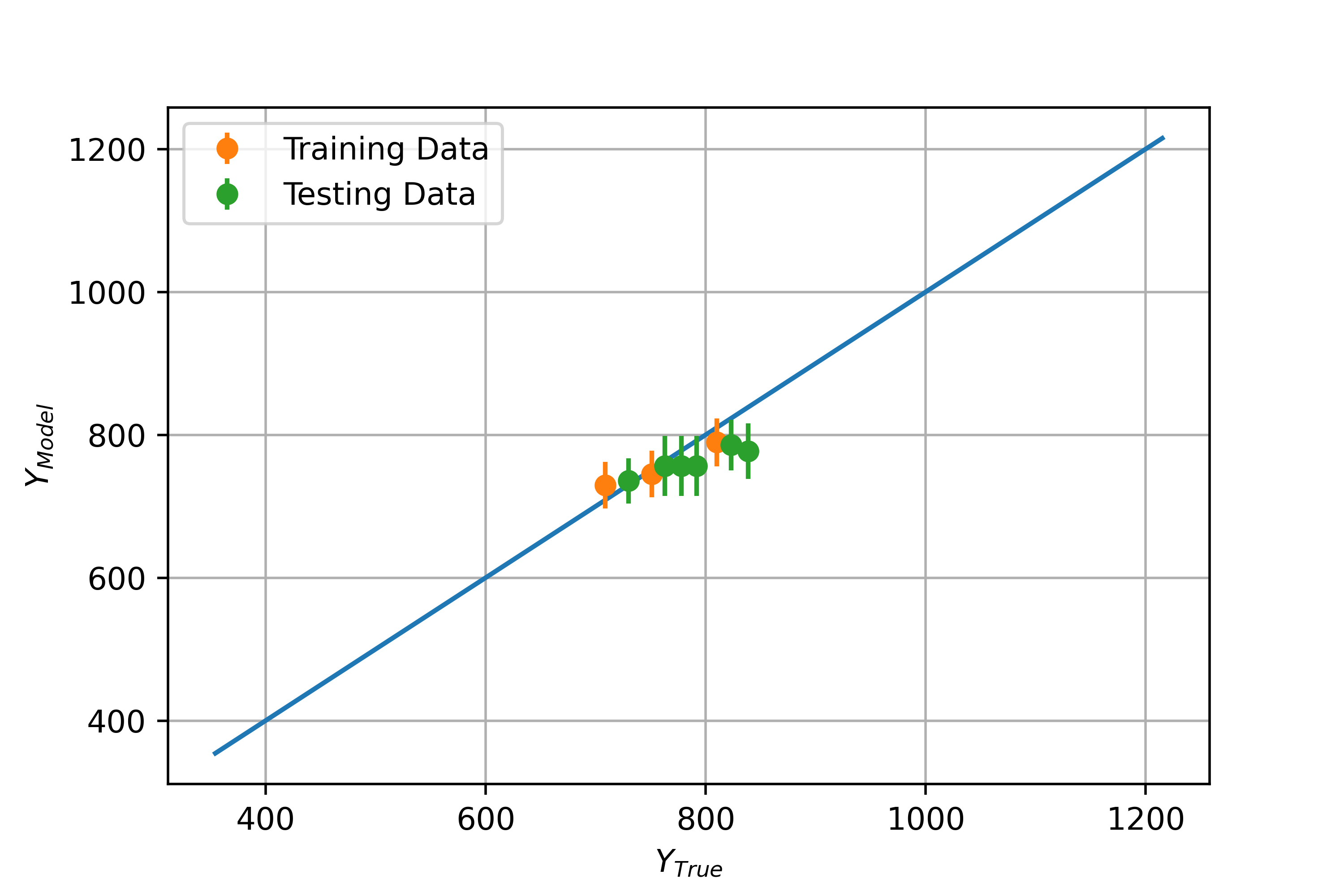}
		\caption{}
		\label{fig:subfig1}
	   \end{subfigure}
	   		       \begin{subfigure}{0.33\linewidth}
		\includegraphics[width=\linewidth]{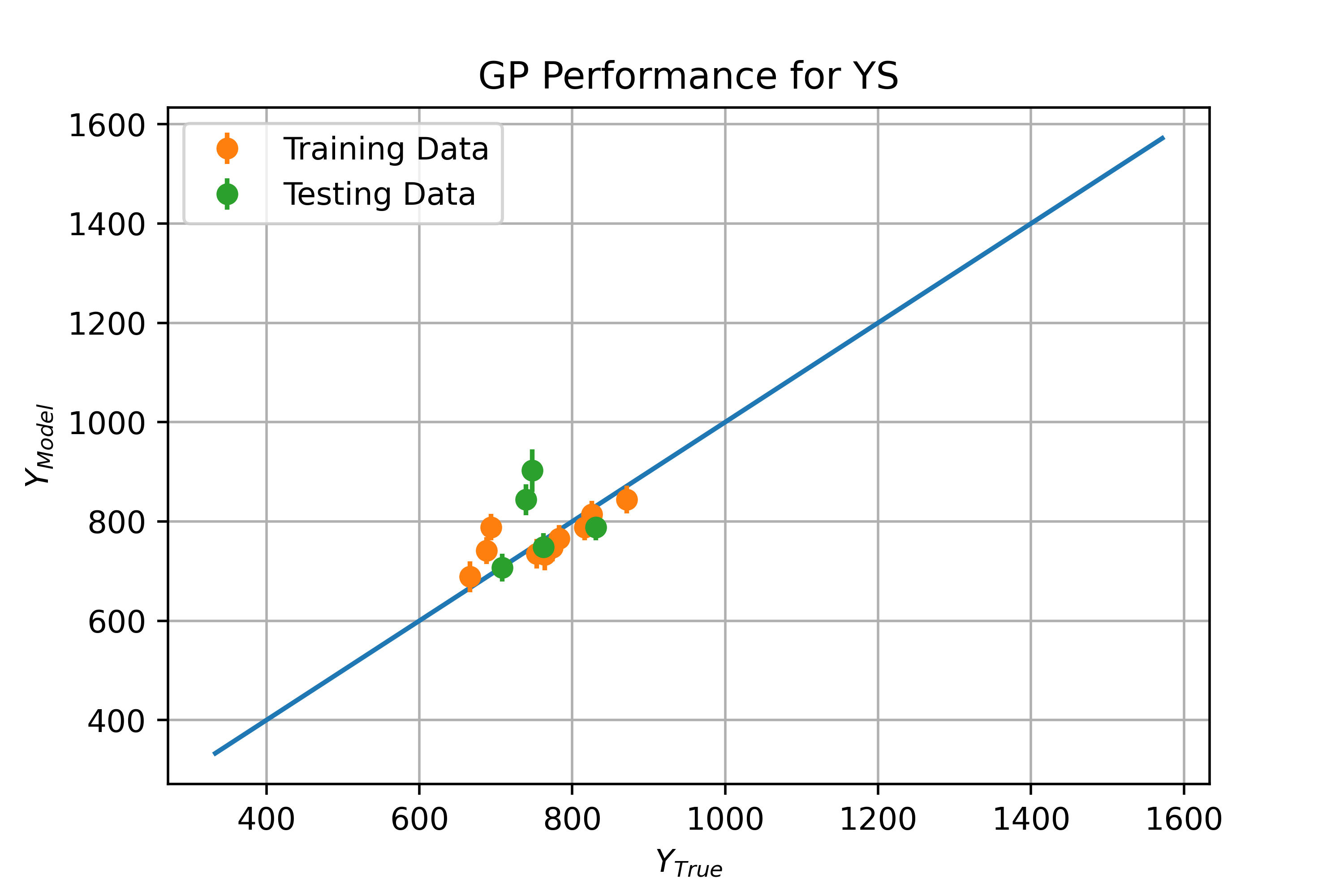}
		\caption{}
		\label{fig:subfig1}
	   \end{subfigure}
	       \begin{subfigure}{0.33\linewidth}
		\includegraphics[width=\linewidth]{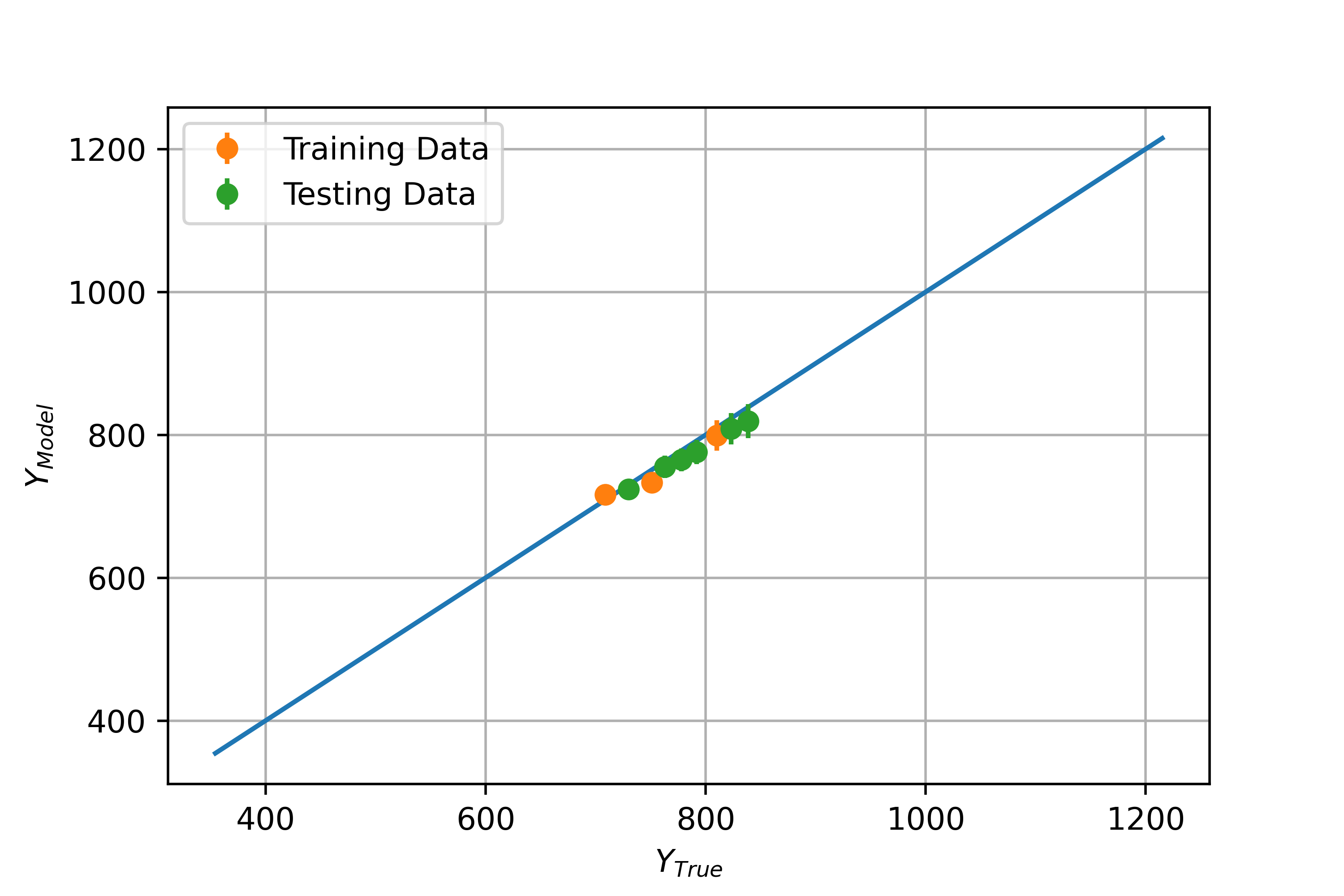}
		\caption{}
		\label{fig:subfig1}
	   \end{subfigure}
	 \vfill
	       \begin{subfigure}{0.45\linewidth}
		  \includegraphics[width=\linewidth]{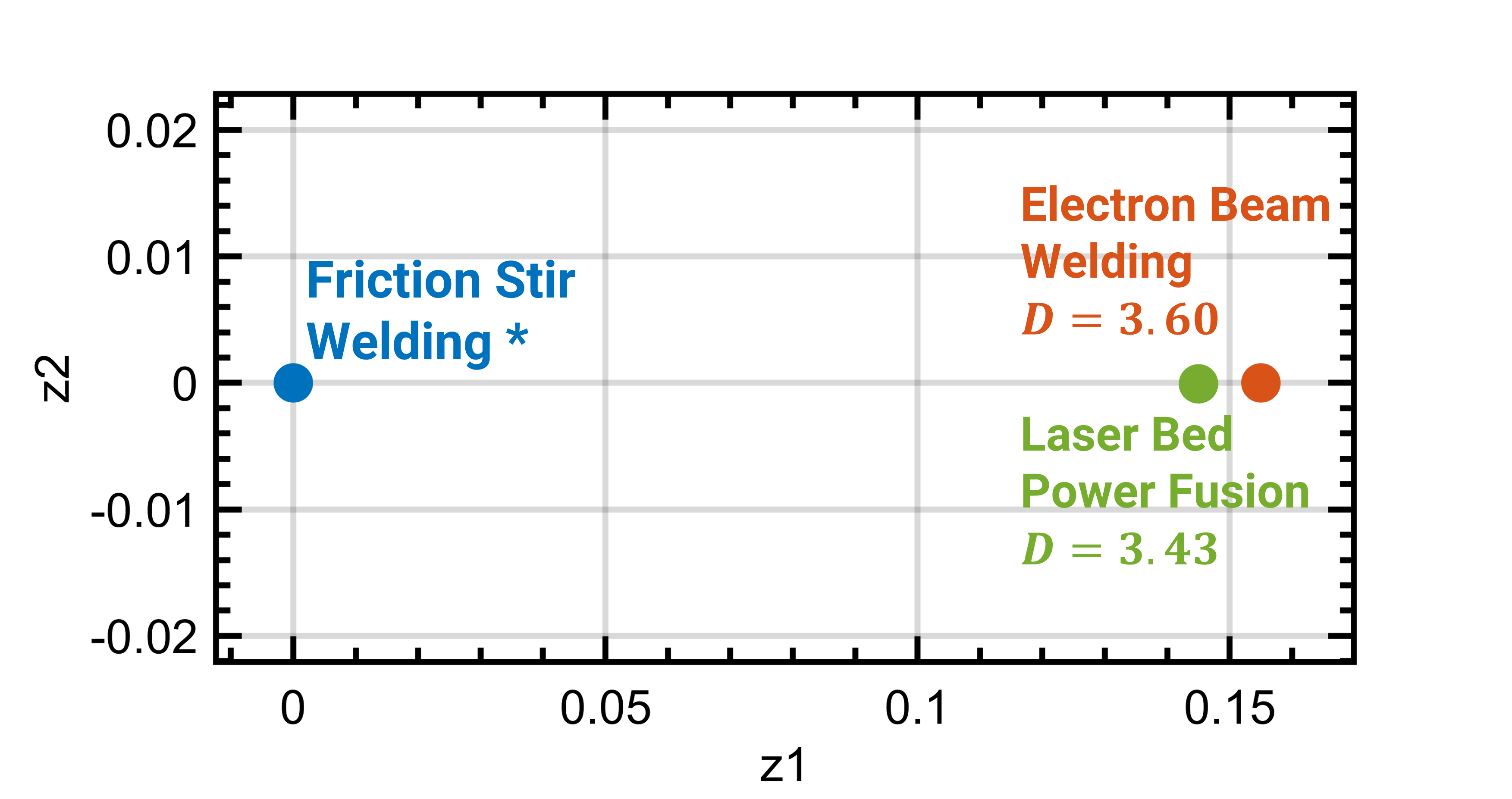}
		  \caption{}
		  \label{fig:subfig4}
	       \end{subfigure}
	\caption{The results of the manufacturing process dusion of Ti6Al4V alloy study. Predictions on (a) all sources using heterogeneous multi-source
GP, (b) sources using heterogeneous multi-source LVGP, (c) friction stir welding using single-source GP built on
original input space, (d) friction stir welding using heterogeneous multi-source GP, (e) friction stir welding using
heterogeneous multi-source LVGP. (f) The latent space of manufacturing modalities obtained by the LVGP model}
	\label{fig:subFigures_Ti6Al4V}
\end{figure}

The true advantage and usefulness of the LVGP approach become apparent when compared with the GP model, built on the original input domain of FSW source data, referred to as GP-FSW in Table \ref{tab:ti_pred}. Due to the limited amount of available data for FSW, this model exhibits limited accuracy on its testing data, as seen in Table \ref{tab:ti_pred}. On the other hand, by transferring knowledge from different manufacturing modalities, the LVGP modeling approach provides significant improvements (of $60.1\%$) in predictions, even though the GP-FSW is built on the non-mapped original input data domain. This stark improvement in the predictions by LVGP can be observed in the parity plots provided by the models as well (Figure \ref{fig:subFigures_Ti6Al4V}c,d\&e). Finally, examining the latent space in Figure \ref{fig:subFigures_Ti6Al4V}f, it is observed that the EBM and LPBF sources are positioned closer together but further from the reference target FSW source. This slight difference between EBM and LPBF could primarily be driven by an underlying hypothesis that the microstructure formation mechanism affecting yield strength is similar due to the high-energy manufacturing processes employed for both modalities.

\clearpage
\section{Conclusion}
Artificial intelligence and machine learning modeling and optimization routines have significantly advanced designs and materials in the engineering community. Amidst this rapid proliferation, a saturation point is often reached concerning the data requirements for creating generalizable and accurate models. To overcome this data deficiency, leveraging different sources of data can provide additional insights into the engineering systems. Each source may vary by several factors, including fidelity, operating conditions, materials, systems, representation of input variables, and more. However, multi-source data fusion methods face challenges when the data sources are heterogeneous, meaning the sources differ by input features with distinct and non-overlapping parameter spaces. To overcome this difficulty, this paper presents a two-stage heterogeneous multi-source data fusion modeling framework. The first stage involves mapping the input feature space of different sources to a common reference source space using the Input Mapping Calibration method. The second stage develops a Latent Variable Gaussian Process based multi-source data fusion model that treats sources as categorical variables, which are later mapped into a low-dimensional latent space that learns the underlying similarities and differences between sources that are not accounted for during the input mapping stage.


The proposed framework was evaluated through three representative engineering case studies with varying challenges: a cantilever beam design problem for sources with varying design representations, an ellipsoidal void design problem for sources with varying fidelities and complexities, and process modeling of an alloy system for sources with different manufacturing modalities that share no common input parameters. These studies demonstrated that the proposed heterogeneous multi-source data fusion modeling framework outperforms conventional multi-source Gaussian process models in terms of predictive capabilities on the transformed space. Furthermore, for the sources with small amount of data, the proposed framework provides significantly better predictive capabilities compared to a single-source Gaussian Process model built solely on the source. Moreover, the framework offers interpretability by identifying underlying similarities and differences between the sources through latent variables and a dissimilarity metric. The proposed method can be further extended to have a broader impact on other research areas including but not limited to digital twins \cite{van2023digital}, multi-task learning \cite{zhang2018overview} where sources are replaced with different tasks, or federated learning \cite{mammen2021federated,li2020review,zhang2021survey} where data are collected on the same process (e.g., additive manufacturing) from different producers, and these data can be integrated using multi-source data fusion by the LVGP. Future research direction of this work can include enhancing the mapping scheme with a non-linear or kernel-based mappings for more accurate frameworks. Additionally, integrating the framework with cost-aware Bayesian optimization for adaptive sampling \cite{CHEN2024116773} could address the challenges of data deficiency and cost associated with AI and ML modeling and design optimization frameworks.

\clearpage
\section*{Acknowledgments}
Northwestern faculty and student are grateful for the sponsorship from NSF 2219489 and the Army Research Laboratory under Cooperative Agreement Number W911NF-22-0121. The views and conclusions contained in this document are those of the authors and should not be interpreted as representing the official policies, either expressed or implied, of the Army Research Laboratory of the U.S. Government. The U.S. Government is authorized to reproduce and distribute reprints for Government purposes notwithstanding any copyright notation herein. This material is based upon work supported by the U.S. Department of Energy’s Office of Energy Efficiency and Renewable Energy (EERE) under the Advanced Manufacturing Office, Award Number DE-AC0206H11357. The views expressed herein do not necessarily represent the views of the U.S. Department of Energy or the United States Government. The authors would also like to express their gratitude and appreciation to Dr. Allison M. Beese and Qixiang Luo for truly embracing the open-data policy and sharing relevant experimental data. The authors would also like to thank Dr. Aymeric Moinet, at GE Aerospace, for his insights into the ellipsoidal void problem.

\section*{Data Availability}
The data used for this work can be requested from the corresponding authors. 


\clearpage

\printbibliography

@inproceedings{krishnan2022transfer,
  title={Transfer Learning Based Modeling of Industrial Turbine Airfoil Characteristics},
  author={Krishnan, S. and Ghosh, S. and Atkinson, S. and Andreoli, V. and Vandeputte, T. and Wang, L.},
  booktitle={AIAA SCITECH 2022 Forum},
  pages={2105},
  year={2022}
}

@inproceedings{ravi2023uncertainty,
  title={On Uncertainty Quantification in Materials Modeling and Discovery: Applications of GE's BHM and IDACE},
  author={Ravi, S.K. and Bhaduri, A. and Amer, A. and Ghosh, S. and Wang, L. and Hoffman, A. and Umretiya, R. and Roy, I. and Rebak, R. and Dheeradhada, V.S. and others},
  booktitle={AIAA SCITECH 2023 Forum}
}

@inproceedings{ravi2023probabilistic,
  title={Probabilistic Transfer Learning through Ensemble Probabilistic Deep Neural Network},
  author={Ravi, S.K. and Pandita, P. and Ghosh, S. and Bhaduri, A. and Andreoli, V. and Wang, L.},
  booktitle={AIAA SCITECH 2023 Forum},
  pages={1479},
  year={2023}
}

@article{comlek2023rapid,
  title={Rapid design of top-performing metal-organic frameworks with qualitative representations of building blocks},
  author={Comlek, Yigitcan and Pham, Thang Duc and Snurr, Randall Q and Chen, Wei},
  journal={npj Computational Materials},
  volume={9},
  number={1},
  pages={170},
  year={2023},
  publisher={Nature Publishing Group UK London}
}

@article{zhang2020latent,
  title={A latent variable approach to Gaussian process modeling with qualitative and quantitative factors},
  author={Zhang, Yichi and Tao, Siyu and Chen, Wei and Apley, Daniel W},
  journal={Technometrics},
  volume={62},
  number={3},
  pages={291--302},
  year={2020},
  publisher={Taylor \& Francis}
}

@article{CHEN2024116773,
title = {A Latent Variable Approach for Non-Hierarchical Multi-Fidelity Adaptive Sampling},
journal = {Computer Methods in Applied Mechanics and Engineering},
volume = {421},
pages = {116773},
year = {2024},
issn = {0045-7825},
doi = {https://doi.org/10.1016/j.cma.2024.116773},
author = {Yi-Ping Chen and Liwei Wang and Yigitcan Comlek and Wei Chen},
}

@article{prabhune2023design,
  title={Design of Polymer Nanodielectrics for Capacitive Energy Storage},
  author={Prabhune, Prajakta and Comlek, Yigitcan and Shandilya, Abhishek and Sundararaman, Ravishankar and Schadler, Linda S and Brinson, Lynda Catherine and Chen, Wei},
  journal={Nanomaterials},
  volume={13},
  number={17},
  pages={2394},
  year={2023},
  publisher={MDPI}
}

@article{iyer2020data,
  title={Data centric nanocomposites design via mixed-variable Bayesian optimization},
  author={Iyer, Akshay and Zhang, Yichi and Prasad, Aditya and Gupta, Praveen and Tao, Siyu and Wang, Yixing and Prabhune, Prajakta and Schadler, Linda S and Brinson, L Catherine and Chen, Wei},
  journal={Molecular Systems Design \& Engineering},
  volume={5},
  number={8},
  pages={1376--1390},
  year={2020},
  publisher={Royal Society of Chemistry}
}

@article{wang2020featureless,
  title={Featureless adaptive optimization accelerates functional electronic materials design},
  author={Wang, Yiqun and Iyer, Akshay and Chen, Wei and Rondinelli, James M},
  journal={Applied Physics Reviews},
  volume={7},
  number={4},
  year={2020},
  publisher={AIP Publishing}
}

@article{taoimc2019,
author = {Tao, Siyu and Apley, Daniel W. and Chen, Wei and Garbo, Andrea and Pate, David J. and German, Brian J.},
title = {Input Mapping for Model Calibration with Application to Wing Aerodynamics},
journal = {AIAA Journal},
volume = {57},
number = {7},
pages = {2734-2745},
year = {2019},
doi = {10.2514/1.J057711},
}

@misc{ravi2024interpretable,
      title={Interpretable Multi-Source Data Fusion Through Latent Variable Gaussian Process}, 
      author={Sandipp Krishnan Ravi and Yigitcan Comlek and Wei Chen and Arjun Pathak and Vipul Gupta and Rajnikant Umretiya and Andrew Hoffman and Ghanshyam Pilania and Piyush Pandita and Sayan Ghosh and Nathaniel Mckeever and Liping Wang},
      year={2024},
      eprint={2402.04146},
      archivePrefix={arXiv},
      primaryClass={stat.ML}
}

@article{tao2019input,
  title={Input mapping for model calibration with application to wing aerodynamics},
  author={Tao, Siyu and Apley, Daniel W and Chen, Wei and Garbo, Andrea and Pate, David J and German, Brian J},
  journal={AIAA journal},
  volume={57},
  number={7},
  pages={2734--2745},
  year={2019},
  publisher={American Institute of Aeronautics and Astronautics}
}

@inproceedings{hebbal2019multi,
  title={Multi-fidelity modeling using DGPs: Improvements and a generalization to varying input space dimensions},
  author={Hebbal, Ali and Brevault, Loic and Balesdent, Mathieu and Talbi, El-Ghazali and Melab, Nouredine},
  booktitle={4th workshop on Bayesian Deep Learning (NeurIPS 2019)},
  year={2019}
}

@article{hebbal2021multi,
  title={Multi-fidelity modeling with different input domain definitions using deep Gaussian processes},
  author={Hebbal, Ali and Brevault, Loic and Balesdent, Mathieu and Talbi, El-Ghazali and Melab, Nouredine},
  journal={Structural and Multidisciplinary Optimization},
  volume={63},
  pages={2267--2288},
  year={2021},
  publisher={Springer}
}

@article{bandler1994space,
  title={Space mapping technique for electromagnetic optimization},
  author={Bandler, John W and Biernacki, Radoslaw M and Chen, Shao Hua and Grobelny, Piotr A and Hemmers, Ronald H},
  journal={IEEE Transactions on microwave theory and techniques},
  volume={42},
  number={12},
  pages={2536--2544},
  year={1994},
  publisher={IEEE}
}

@article{koziel2018implicit,
  title={Implicit space mapping for variable-fidelity EM-driven design of compact circuits},
  author={Koziel, S and Bekasiewicz, Adrian},
  journal={IEEE Microwave and Wireless Components Letters},
  volume={28},
  number={4},
  pages={275--277},
  year={2018},
  publisher={IEEE}
}

@article{jiang2018space,
  title={A space mapping method based on Gaussian process model for variable fidelity metamodeling},
  author={Jiang, Ping and Xie, Tingli and Zhou, Qi and Shao, Xinyu and Hu, Jiexiang and Cao, Longchao},
  journal={Simulation Modelling Practice and Theory},
  volume={81},
  pages={64--84},
  year={2018},
  publisher={Elsevier}
}

@article{day2017survey,
  title={A survey on heterogeneous transfer learning},
  author={Day, Oscar and Khoshgoftaar, Taghi M},
  journal={Journal of Big Data},
  volume={4},
  pages={1--42},
  year={2017},
  publisher={Springer}
}

@article{bao2023survey,
  title={A Survey on Heterogeneous Transfer Learning},
  author={Bao, Runxue and Sun, Yiming and Gao, Yuhe and Wang, Jindong and Yang, Qiang and Chen, Haifeng and Mao, Zhi-Hong and Xie, Xing and Ye, Ye},
  journal={arXiv preprint arXiv:2310.08459},
  year={2023}
}

@inproceedings{shi2010transfer,
  title={Transfer learning on heterogenous feature spaces via spectral transformation},
  author={Shi, Xiaoxiao and Liu, Qi and Fan, Wei and Philip, S Yu and Zhu, Ruixin},
  booktitle={2010 IEEE international conference on data mining},
  pages={1049--1054},
  year={2010},
  organization={IEEE}
}

@article{liu2023learning,
  title={Learning Multitask Gaussian Process Over Heterogeneous Input Domains},
  author={Liu, Haitao and Wu, Kai and Ong, Yew-Soon and Bian, Chao and Jiang, Xiaomo and Wang, Xiaofang},
  journal={IEEE Transactions on Systems, Man, and Cybernetics: Systems},
  year={2023},
  publisher={IEEE}
}

@article{gorodetsky2020mfnets,
  title={MFNets: Multi-fidelity data-driven networks for Bayesian learning and prediction},
  author={Gorodetsky, Alex A and Jakeman, John D and Geraci, Gianluca and Eldred, Michael S},
  journal={International Journal for Uncertainty Quantification},
  volume={10},
  number={6},
  year={2020},
  publisher={Begel House Inc.}
}

@article{jin2021combining,
  title={Combining point and distributed strain sensor for complementary data-fusion: A multi-fidelity approach},
  author={Jin, Seung-Seop and Kim, Sung Tae and Park, Young-Hwan},
  journal={Mechanical Systems and Signal Processing},
  volume={157},
  pages={107725},
  year={2021},
  publisher={Elsevier}
}

@article{duan2012learning,
  title={Learning with augmented features for heterogeneous domain adaptation},
  author={Duan, Lixin and Xu, Dong and Tsang, Ivor},
  journal={arXiv preprint arXiv:1206.4660},
  year={2012}
}

@inproceedings{wang2011heterogeneous,
  title={Heterogeneous domain adaptation using manifold alignment},
  author={Wang, Chang and Mahadevan, Sridhar},
  booktitle={IJCAI proceedings-international joint conference on artificial intelligence},
  volume={22},
  number={1},
  pages={1541},
  year={2011}
}

@article{yousefpour2024gp+,
  title={GP+: a python library for kernel-based learning via Gaussian Processes},
  author={Yousefpour, Amin and Foumani, Zahra Zanjani and Shishehbor, Mehdi and Mora, Carlos and Bostanabad, Ramin},
  journal={Advances in Engineering Software},
  volume={195},
  pages={103686},
  year={2024},
  publisher={Elsevier}
}

@article{eweis2022data,
  title={Data fusion with latent map Gaussian processes},
  author={Eweis-Labolle, Jonathan Tammer and Oune, Nicholas and Bostanabad, Ramin},
  journal={Journal of Mechanical Design},
  volume={144},
  number={9},
  pages={091703},
  year={2022},
  publisher={American Society of Mechanical Engineers}
}

@article{foumani2023multi,
  title={Multi-fidelity cost-aware Bayesian optimization},
  author={Foumani, Zahra Zanjani and Shishehbor, Mehdi and Yousefpour, Amin and Bostanabad, Ramin},
  journal={Computer Methods in Applied Mechanics and Engineering},
  volume={407},
  pages={115937},
  year={2023},
  publisher={Elsevier}
}

@article{zanjani2024safeguarding,
  title={Safeguarding Multi-fidelity Bayesian Optimization Against Large Model Form Errors and Heterogeneous Noise},
  author={Zanjani Foumani, Zahra and Yousefpour, Amin and Shishehbor, Mehdi and Bostanabad, Ramin},
  journal={Journal of Mechanical Design},
  volume={146},
  number={6},
  year={2024},
  publisher={American Society of Mechanical Engineers Digital Collection}
}

@article{menon2024multi,
  title={Multi-fidelity surrogate with heterogeneous input spaces for modeling melt pools in laser-directed energy deposition},
  author={Menon, Nandana and Basak, Amrita},
  journal={arXiv preprint arXiv:2403.13136},
  year={2024}
}

@article{ran2020microstructure,
  title={Microstructure and mechanical properties of Ti-6Al-4V fabricated by electron beam melting},
  author={Ran, Jiangtao and Jiang, Fengchun and Sun, Xiaojing and Chen, Zhuo and Tian, Cao and Zhao, Hong},
  journal={Crystals},
  volume={10},
  number={11},
  pages={972},
  year={2020},
  publisher={MDPI}
}

@article{fall2017effect,
  title={Effect of process parameters on microstructure and mechanical properties of friction stir-welded Ti--6Al--4V joints},
  author={Fall, A and Jahazi, M and Khdabandeh, AR and Fesharaki, MH},
  journal={The International Journal of Advanced Manufacturing Technology},
  volume={91},
  pages={2919--2931},
  year={2017},
  publisher={Springer}
}

@article{luo2022effect,
  title={Effect of processing parameters on pore structures, grain features, and mechanical properties in Ti-6Al-4V by laser powder bed fusion},
  author={Luo, Qixiang and Yin, Lu and Simpson, Timothy W and Beese, Allison M},
  journal={Additive Manufacturing},
  volume={56},
  pages={102915},
  year={2022},
  publisher={Elsevier}
}

@article{luo2023dataset,
  title={Dataset of process-structure-property feature relationship for laser powder bed fusion additive manufactured Ti-6Al-4V material.},
  author={Luo, Qixiang and Yin, Lu and Simpson, Timothy W and Beese, Allison M},
  journal={Data in Brief},
  volume={46},
  pages={108911},
  year={2023},
  publisher={Elsevier}
}

@article{pan2010domain,
  title={Domain adaptation via transfer component analysis},
  author={Pan, Sinno Jialin and Tsang, Ivor W and Kwok, James T and Yang, Qiang},
  journal={IEEE transactions on neural networks},
  volume={22},
  number={2},
  pages={199--210},
  year={2010},
  publisher={IEEE}
}

@article{wang2021data,
  title={Data-driven topology optimization with multiclass microstructures using latent variable Gaussian process},
  author={Wang, Liwei and Tao, Siyu and Zhu, Ping and Chen, Wei},
  journal={Journal of Mechanical Design},
  volume={143},
  number={3},
  pages={031708},
  year={2021},
  publisher={American Society of Mechanical Engineers}
}

@article{karniadakis2021physics,
  title={Physics-informed machine learning},
  author={Karniadakis, George Em and Kevrekidis, Ioannis G and Lu, Lu and Perdikaris, Paris and Wang, Sifan and Yang, Liu},
  journal={Nature Reviews Physics},
  volume={3},
  number={6},
  pages={422--440},
  year={2021},
  publisher={Nature Publishing Group}
}

@article{cuomo2022scientific,
  title={Scientific machine learning through physics--informed neural networks: Where we are and what’s next},
  author={Cuomo, Salvatore and Di Cola, Vincenzo Schiano and Giampaolo, Fabio and Rozza, Gianluigi and Raissi, Maziar and Piccialli, Francesco},
  journal={Journal of Scientific Computing},
  volume={92},
  number={3},
  pages={88},
  year={2022},
  publisher={Springer}
}

@inproceedings{luan2024physics,
  title={Physics-Informed Research Assistant for Theory Extraction (PIRATE) for Missing Physics Discovery},
  author={Luan, Lele and Jacobs, Ryan and Ghosh, Sayan and Wang, Liping},
  booktitle={AIAA SCITECH 2024 Forum},
  pages={0171},
  year={2024}
}

@inproceedings{luan2023physics,
  title={Physics Discovery of Engineering Applications With Constrained Optimization and Genetic Programming},
  author={Luan, Lele and Jacobs, Ryan and Ghosh, Sayan and Wang, Liping},
  booktitle={Turbo Expo: Power for Land, Sea, and Air},
  volume={87066},
  pages={V11BT25A005},
  year={2023},
  organization={American Society of Mechanical Engineers}
}

@article{zhang2018overview,
  title={An overview of multi-task learning},
  author={Zhang, Yu and Yang, Qiang},
  journal={National Science Review},
  volume={5},
  number={1},
  pages={30--43},
  year={2018},
  publisher={Oxford University Press}
}

@article{li2020review,
  title={A review of applications in federated learning},
  author={Li, Li and Fan, Yuxi and Tse, Mike and Lin, Kuo-Yi},
  journal={Computers \& Industrial Engineering},
  volume={149},
  pages={106854},
  year={2020},
  publisher={Elsevier}
}

@article{mammen2021federated,
  title={Federated learning: Opportunities and challenges},
  author={Mammen, Priyanka Mary},
  journal={arXiv preprint arXiv:2101.05428},
  year={2021}
}

@article{zhang2021survey,
  title={A survey on federated learning},
  author={Zhang, Chen and Xie, Yu and Bai, Hang and Yu, Bin and Li, Weihong and Gao, Yuan},
  journal={Knowledge-Based Systems},
  volume={216},
  pages={106775},
  year={2021},
  publisher={Elsevier}
}

@article{van2023digital,
  title={Digital twins for the designs of systems: a perspective},
  author={van Beek, Anton and Nevile Karkaria, Vispi and Chen, Wei},
  journal={Structural and Multidisciplinary Optimization},
  volume={66},
  number={3},
  pages={49},
  year={2023},
  publisher={Springer}
}

\clearpage
\setcounter{equation}{0}
\setcounter{figure}{0}
\setcounter{table}{0}


\end{document}